\documentclass[11pt]{article}

\usepackage[preprint]{acl}

\usepackage{times}
\usepackage{latexsym}

\usepackage[T1]{fontenc}
\usepackage[utf8]{inputenc}

\usepackage{microtype}

\usepackage{inconsolata}

\usepackage{graphicx}

\usepackage{amsmath, amssymb, amsfonts}
\usepackage{bm}
\usepackage{dsfont}
\usepackage{latexsym}
\usepackage{algorithmic}
\usepackage{algorithm}

\usepackage{graphicx}
\usepackage{subcaption}
\usepackage{booktabs}
\usepackage{longtable}
\usepackage{array}
\usepackage{changepage}
\usepackage{multirow}
\usepackage{makecell}
\usepackage{siunitx}
\usepackage[table]{xcolor}
\usepackage{caption}
\usepackage{tcolorbox}
\tcbuselibrary{skins,breakable}

\definecolor{lightteal}{RGB}{234,209,220}
\definecolor{lightblue}{RGB}{181, 179, 242}
\definecolor{teal}{RGB}{116,27,71}
\definecolor{promptcolor}{RGB}{70, 70, 70}
\definecolor{examplecolor}{RGB}{45, 85, 135}
\definecolor{darkgreen}{RGB}{0, 100, 0}

\usepackage{listings}
\usepackage{amssymb}
\usepackage{pifont}

\newcommand{\dec}[2]{\shortstack{#1\\[-1pt]\tiny\textcolor{red}{$\downarrow$#2}}}
\newtcolorbox[auto counter, number within=section]{prompt}[2][]{
    title={Prompt \thetcbcounter: #2},
    colback=white,
    colbacktitle=promptcolor,
    coltitle=white,
    fonttitle=\bfseries\small\sffamily,
    fontupper=\small\sffamily,
    top=8pt,
    bottom=8pt,
    left=8pt,
    right=8pt,
    boxrule=1pt,
    breakable,
    #1
}

\newtcolorbox[auto counter, number within=section]{example}[2][]{
    title={Example \thetcbcounter: #2},
    colback=lightblue!10, 
    colbacktitle=examplecolor,
    coltitle=white,
    fonttitle=\bfseries\small\sffamily,
    fontupper=\scriptsize\sffamily,
    top=1pt,
    bottom=1pt,
    left=1pt,
    right=1pt,
    boxrule=1pt,
    breakable,
    before upper={\scriptsize\sffamily},
    #1
}

\title{Decompose-and-Formalise: \\Recursively Verifiable Natural Language Inference}

\author{Xin Quan$^1$, Marco Valentino$^{2}$,  Louise A. Dennis$^1$, Andr\'e Freitas$^{1,3,4}$ \\ 
$^{1}$Department of Computer Science, University of Manchester, UK \\ 
$^{2}$School of Computer Science, University of Sheffield, UK \\ 
$^{3}$Idiap Research Institute, Switzerland \\
$^{4}$National Biomarker Centre, CRUK-MI, University of Manchester, UK\\
  \texttt{xin.quan@manchester.ac.uk}\quad \texttt{m.valentino@sheffield.ac.uk}\\
  \texttt{louise.dennis@manchester.ac.uk}\quad \texttt{andre.freitas@idiap.ch}
}

\begin{document}
\maketitle
\begin{abstract}
%Natural language explanations are central to human reasoning, yet they are often incomplete and logically fragile, motivating methods that make their semantic commitments explicit and refine them into defensible stepwise reasoning. 
Recent work has shown that integrating large language models (LLMs) with theorem provers (TPs) in neuro-symbolic pipelines helps with entailment verification and proof-guided refinement of explanations for natural language inference (NLI). However, scaling such refinement to naturalistic NLI remains difficult: long, syntactically rich inputs and deep multi-step arguments amplify autoformalisation errors, where a single local mismatch can invalidate the proof. Moreover, current methods often handle  failures via costly global regeneration due to the difficulty of localising the responsible span or step from prover diagnostics. Aiming to address these problems, we propose a decompose-and-formalise framework that (i) decomposes premise-hypothesis pairs into an entailment tree of atomic steps, (ii) verifies the tree bottom-up to isolate failures to specific nodes, and (iii) performs local diagnostic-guided refinement instead of regenerating the whole explanation. Moreover, to improve faithfulness of autoformalisation, we introduce $\theta$-substitution in an event-based logical form to enforce consistent argument–role bindings. Across a range of reasoning tasks using five LLM backbones, our method achieves the highest explanation verification rates, improving over the state-of-the-art by 26.2\%, 21.7\%, 21.6\% and 48.9\%, while reducing refinement iterations and runtime and preserving strong NLI accuracy.
\end{abstract}

\begin{figure*}[t]
    \centering
    \includegraphics[width=\textwidth]{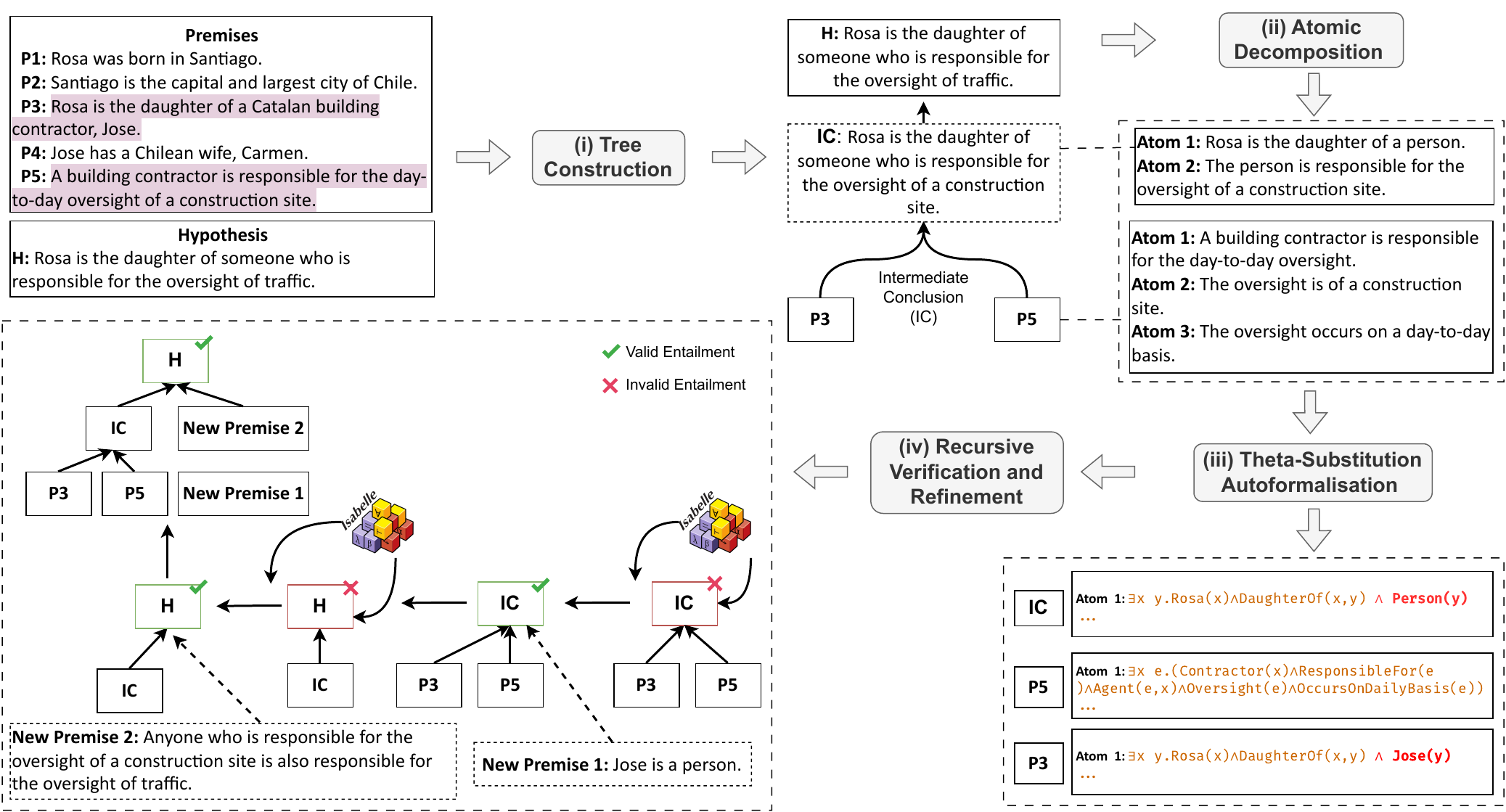}
    \caption{An illustration of the proposed framework for recursively verifiable natural language inference (NLI) via entailment trees, atomic decomposition, and $\theta$-substitution autoformalisation. The entailment tree is initially constructed from the given premises and hypothesis, including intermediate conclusions. Each sentence is then decomposed into atomic propositions and autoformalised into logical forms. By utilising an external theorem prover, we progressively verify and refine each subtree until the final hypothesis node is reached. The reasoning chain is considered logically sound and coherent once the entire entailment tree has been fully verified and refined.}
\label{fig:framework}
\end{figure*}

\section{Introduction}

Natural language explanations are the workhorse of human communication and rationality, yet they are frequently incomplete, imprecise, and logically fragile \cite{valentino2024nature,valentino2021natural}. This motivates verifiable Natural Language Inference (NLI) systems that can make the implicit commitments of everyday explanations explicit, and refine them into solver-checkable stepwise certificates~\citep{quan-etal-2024-verification, sadeddine-suchanek-2025-verifying}. Recent works in neuro-symbolic pipelines offer a principled route to this goal by integrating Large Language Models (LLMs) with external theorem provers (TPs): LLMs autoformalise natural language statements into formal representations, and TPs then verify whether premises and explanations entail a hypothesis in the target logic~\citep{pan-etal-2023-logic, olausson-etal-2023-linc, quan-etal-2025-peirce,NEURIPS2022_d0c6bc64, xu-etal-2024-faithful, xu2025adaptivellmsymbolicreasoningdynamic, zhang-etal-2025-autoformalization, jiang-etal-2024-leanreasoner}. TP feedback can further support iterative refinement of both formalisation and explanations~\citep{quan-etal-2024-verification, quan-etal-2025-faithful, xu2025trainingllmslogicrewardfaithful}.

Despite the pace of progress, scaling formal refinement to naturalistic and complex NLI remains an open problem. Real explanations often involve long, syntactically rich sentences and deep multi-step arguments that compress many coupled semantic commitments into a small surface form. A single local mismatch (such as a role swap, a negation-scope error, or a quantifier mistake) can derail an entire reasoning chain. Moreover, because these errors propagate across steps, end-to-end regeneration and re-proving becomes increasingly unstable and costly as chains become longer and more branched~\citep{pan-etal-2023-logic, quan-etal-2025-faithful}. At the same time, autoformalisation must be both prover-compatible and semantically faithful (preserving scope, roles, quantification, and lexical commitments) because even minor semantic drift can yield proofs that are syntactically valid but misaligned with the original text~\citep{jiang-etal-2024-leanreasoner, xu2025trainingllmslogicrewardfaithful, quan-etal-2025-faithful}. These bottlenecks foreground three questions: \textit{RQ1: How can we obtain autoformalisation that is both prover-compatible and semantically faithful?} \textit{RQ2: What kind of verification strategy provides transparent, stepwise verification that preserves multi-step structure and reduces refinement drift?} \textit{RQ3: To what extent can local diagnostic-guided refinement improve efficiency and robustness without sacrificing end-task accuracy?}

To address these challenges, we propose \textit{LLM-TP Tree}, a scalable decompose-and-formalise framework built around an entailment-tree view of explanations. The core idea is to replace holistic refinement with structured decomposition and local repair. Concretely, LLM-TP Tree (i) decomposes each premise–hypothesis instance into an entailment tree of atomic steps, (ii) verifies the tree bottom-up to isolate proof failures to specific nodes, and (iii) performs local diagnostic-guided refinement on the failing nodes rather than regenerating the entire explanation. This design makes semantic preservation controllable at the smallest meaningful granularity, stabilises refinement under longer-horizon reasoning, and improves efficiency by focusing computation where proofs actually fail.
%A central obstacle to faithful autoformalisation in this setting is maintaining consistency of argument-role bindings across steps and across repeated repairs. 
To improve faithfulness in autoformalisation, we introduce a $\theta$-substitution procedure within an event-based logical form that enforces consistent argument-role bindings during autoformalisation, making role commitments explicit and systematically repairable. Together, atomic decomposition, recursive bottom-up verification, and $\theta$-substitution yield prover-compatible theories that better track the intended semantics of the original text while enabling targeted refinement.

We evaluate LLM-TP Tree on FOLIO, ProofWriter, PrOntoQA, and EntailmentBank using five LLM backbones (GPT-4o, GPT-5 nano, Grok-4-fast, Deepseek-V3.1 and Qwen3-Max). We compare against state-of-the-art baselines including Explanation-Refiner~\citep{quan-etal-2024-verification}, Faithful-Refiner~\citep{quan-etal-2025-faithful}, LINC~\citep{olausson-etal-2023-linc}, and Logic-LM~\citep{pan-etal-2023-logic}. Across datasets, our approach achieves the strongest refinement performance, improving the average verification rate by 26.2\%, 21.7\%, 21.6\%, and 48.9\%  points % to 87.70\%, 94.53\%, 97.33\%, and 72.53\%
on FOLIO, ProofWriter, PrOntoQA, and EntailmentBank, respectively, while reducing refinement iterations and runtime and maintaining strong NLI accuracy. In summary, our contributions are:
\begin{enumerate}
\item We introduce LLM-TP Tree, a recursive entailment-tree framework that verifies and refines multi-step inference chains for verifiable NLI via bottom-up checking and local repair.
\item We propose atomic decomposition, which decomposes complex sentences into atomic propositions and reduces inference-time by 16.5\%, 13.7\%, 21.1\%, and 10.1\% across datasets compared to baseline neuro-symbolic frameworks.
\item We propose a $\theta$-substitution autoformalisation procedure that enforces consistent argument-role bindings and improves autoformalisation faithfulness by 0.09, 0.106, 0.128, and 0.21 on average over baseline models.
\item We conduct comprehensive automatic and human evaluations with ablations, showing that LLM-TP Tree substantially improves the robustness and efficiency of LLM-TP explanation refinement while preserving strong logical reasoning performance.
\end{enumerate}

\section{Recursively Verifiable NLI}
In this work, we define an NLI instance $i$ with a premise set $P_i=\{p_1,p_2,\dots,p_n\}$ and a hypothesis $h_i$ to be formally verifiable if the entailment condition can be verified by a theorem prover. Let $\Phi:\mathrm{NL}\to\mathcal{L}$ translate natural language sentences into a target logic $\mathcal{L}$, and let $\mathcal{S}$ be a solver for $\mathcal{L}$. The NLI instance $i$ is labelled as entailment if $\mathcal{S}(\Phi(P_i)\vdash \Phi(h_i))$ is provable, and as contradiction if $\mathcal{S}(\Phi(P_i)\vdash \neg\Phi(h_i))$ is provable.

Formal verifiability certifies only the endpoint relation between $P_i$ and $h_i$. We call an instance recursively verifiable if there exists an explicit inference chain $\Pi=\langle \delta_1,\delta_2,\dots,\delta_k\rangle$ whose intermediate steps are individually
solver-checkable and compose to $h_i$. Let $\Pi^{<j}=\langle \delta_1,\dots,\delta_{j-1}\rangle$ denote the prefix before step $j$. Then $\Pi$ is a recursively verifiable witness if, for all $j\in\{1,\dots,k\}$, $\mathcal{S}\big(\Phi(P_i\cup \Pi^{<j})\vdash \Phi(\delta_j)\big)$ is provable, and $\mathcal{S}\big(\Phi(P_i\cup \Pi)\vdash \Phi(h_i)\big)$ is provable.

%To connect the chain-based definition above with a tree-structured reasoning witness used in the following sections, we show that the local obligations in an entailment tree are equivalent to those in an ordered chain.

%\begin{theorem}[Tree--Sequent Equivalence]
% \label{thm:tree-sequent}
%Let $T$ be an entailment tree whose internal nodes are $C=\{\delta_1,\ldots,\delta_k\}$ and leaves in
%$\Gamma$. If every internal node $v$ satisfies $\bigwedge_{u\in\mathcal{D}(v)}\varphi_u \vdash \varphi_v$
%validated by $\mathcal{S}$, then there exists an ordering of $C$ forming $\Pi$ such that
%$\forall j:\ \Gamma,\Pi^{<j}\vdash \delta_j$ and $\Gamma,\Pi\vdash h$. Conversely, any such $\Pi$ induces
%a closed tree $T$.
%\end{theorem}

\section{Methodology}
To enable faithful formal checking of multi-step reasoning chains and to refine them when proof obligations are not discharged, we propose LLM-TP Tree, a neuro symbolic framework for recursively verifiable NLI based on an entailment reasoning tree $T_i$. Figure~\ref{fig:framework} provides an overview.

\subsection{Entailment Tree Construction}
Given a premise set $P_i$ and a hypothesis $h_i$, we first prompt an LLM to construct an entailment tree and a set of intermediate conclusions
$C=\{\delta_1,\delta_2,\dots,\delta_m\}$, optionally selecting a subset of premises $P'_i\subseteq P_i$, such that $P'_i\cup C \models h_i$ holds. The tree $T_i$ consists of nodes labelled by propositions (premises or intermediate conclusions) and directed edges encoding entailment dependencies.
For each internal node $v\in T_i$ with proposition $\varphi_v$ and descendant nodes $\mathcal{D}(v)$, the corresponding local entailment obligation requires that the conjunction of descendants entails the parent, namely $\bigwedge_{u\in\mathcal{D}(v)} \varphi_u \models \varphi_v$. The prompt of the entailment tree construction is in Appendix \ref{appendix:entailment_tree}.

\subsection{Atomic Decomposition}
Sentence complexity is a major contributor to both the latency and the brittleness of
autoformalisation \citep{quan-etal-2025-faithful}. Long and structurally complex sentences tend to increase the search space of the translation process, which not only slows down inference but also amplifies semantic inconsistencies in the resulting logical forms. To mitigate these issues, we then decompose each natural language proposition into a set of atomic propositions, so that the subsequent autoformalisation operates on shorter, semantically focused units.

After constructing the entailment reasoning tree, we apply atomic decomposition to the propositions (first subtree) that participate in proof obligations. For the hypothesis $h_i$, we generate a set of atoms
\begin{equation}
D(h_i)=\{a^{h}_{1}, a^{h}_{2}, \ldots, a^{h}_{k}\},
\qquad
h_i \equiv \bigwedge_{\ell=1}^{k} a^{h}_{\ell}.
\end{equation}

Similarly, for the premise set $P_i=\{p_1,\ldots,p_n\}$, each premise $p_j$ is decomposed into

\begin{equation}
D(p_j)=\{a^{p_j}_{1}, a^{p_j}_{2}, \ldots, a^{p_j}_{t_j}\},
\qquad
p_j \equiv \bigwedge_{\ell=1}^{t_j} a^{p_j}_{\ell}.
\end{equation}

We define the global atom set induced by the premises as

\begin{equation}
D(P_i) := \bigcup_{j=1}^{n} D(p_j),
\qquad
P_i \equiv \bigwedge_{a\in D(P_i)} a.
\end{equation}

\paragraph{Entailment-preserving requirement.}
We require the decomposition operator $D(\cdot)$ to be entailment-preserving: for any sentence $\varphi$ and any atom $a\in D(\varphi)$, it must hold that $\varphi \vDash a$. Intuitively, each atom is a logically entailed consequence of the original sentence, so decomposition does not introduce new information that is not supported by the source. We obtain candidate atoms by prompting an LLM (see Appendix \ref{appendix:atomic_decompistion_prompt}). To enforce the entailment-preserving requirement in practice, we apply a pretrained NLI classifier
(\texttt{roberta-large-mnli}~\citep{liu2019roberta}) to verify that each atom is entailed by its source sentence within a predefined threshold (0.9). This filtering step reduces invalid decompositions and improves the stability of downstream autoformalisation.

\begin{figure*}[t]
\centering
    \begin{example}[breakable=false]{
    \texorpdfstring{$\theta$}{theta}-substitution Autoformalisation}
    \textbf{Sentence:} A forest fire would cause deer to die or leave a woodland.
    \vspace{0.5em} 
    
    \textbf{Logic Template Generation:}
    \begin{equation}
    \varphi_0 = \forall x\ y\ z.\ P(x) \land Q(y) \land R(z) \rightarrow S
    \end{equation}
    
    \textbf{Step 1: Entity Substitution} (Apply $\theta_1 = \{P/\text{ForestFire}, Q/\text{Deer}, R/\text{Woodland}\}$):
    \begin{equation}
    \varphi_1 = \varphi_0[\theta_1] = \forall x\ y\ z.\ \text{ForestFire}(x) \land \text{Deer}(y) \land \text{Woodland}(z) \rightarrow S
    \end{equation}
    
    \textbf{Step 2: Event Substitution} (Apply $\theta_2 = \{S/(\text{Die}(e_1) \lor \text{Leave}(e_2))\}$):
    \begin{equation}
    \varphi_2 = \varphi_1[\theta_2] = \forall x\ y\ z.\ \text{ForestFire}(x) \land \text{Deer}(y) \land \text{Woodland}(z) \rightarrow (\text{Die}(e_1) \lor \text{Leave}(e_2))
    \end{equation}
    
    \textbf{Step 3: Semantic Role Substitution} (Apply $\theta_3$):
    \begin{equation}
    \theta_3 = \begin{cases}
    \text{Die}(e_1) \mapsto (\text{Die}(e_1) \land \text{Agent}(e_1, y)) \\
    \text{Leave}(e_2) \mapsto (\text{Leave}(e_2) \land \text{Agent}(e_2, y) \land \text{Patient}(e_2, z))
    \end{cases}
    \end{equation}
    
    \textbf{Final Result:}
    \begin{equation}
    \begin{split} 
        \varphi_{\text{final}} &= \varphi_2[\theta_3] = \forall x\ y\ z\ e_1\ e_2.\ \text{ForestFire}(x) \land \text{Deer}(y) \land \text{Woodland}(z) \rightarrow \\
        &\quad (\text{Die}(e_1) \land \text{Agent}(e_1, y)) \lor (\text{Leave}(e_2) \land \text{Agent}(e_2, y) \land \text{Patient}(e_2, z))
    \end{split}
    \end{equation}
  \end{example}
  \caption{$\theta$-substitution autoformalisation example.}
  \label{fig:theta_example}
\end{figure*}

\subsection{\texorpdfstring{$\theta$}{theta}-substitution Autoformalisation}
Existing autoformalisation pipelines~\citep{pan-etal-2023-logic, olausson-etal-2023-linc} often rely on complex, monolithic generation procedures to map natural language sentences into fully specified logical formulas. This design is computationally inefficient and, more importantly, prone to systematic errors that frequently prevent theorem provers from certifying entailment. These observations motivate a more structured autoformalisation strategy that decomposes the translation into smaller, verifiable steps and reduces sensitivity to surface linguistic variation.

We then adopt the Neo-Davidsonian event semantics formulation used in \citet{quan-etal-2024-verification}, where events and semantic roles (e.g., agent, patient) are represented explicitly in first-order logic to autoformalise the decomposed atomic propositions in previous step to logical forms. Building on this representation, we propose a multi-step $\theta$-substitution procedure that constructs the final logical form by progressively instantiating an abstract template (see Example~\ref{fig:theta_example}).

A substitution $\theta$ is a mapping from placeholders to terms:

\begin{equation}
\theta = \{x_1 \mapsto t_1,\; x_2 \mapsto t_2,\; \ldots,\; x_n \mapsto t_n\}.
\end{equation}

Applying $\theta$ to a formula $\varphi$, denoted $\varphi[\theta]$, replaces each placeholder $x_i$ with its corresponding term $t_i$.

For a natural language sentence, we start from an abstract logical template and apply a sequence of substitutions:
\begin{equation}
\varphi_0 \xrightarrow{\theta_1} \varphi_1 \xrightarrow{\theta_2} \varphi_2 \xrightarrow{\theta_3} \varphi_{\text{final}}.
\end{equation}

Each substitution handles a specific aspect: 1) \textit{Entity substitution} ($\theta_1$): maps generic predicates to specific entities 2) \textit{Event substitution} ($\theta_2$): introduces event predicates and event structure. 3) \textit{Role substitution} ($\theta_3$): adds Neo-Davidsonian semantic roles and their arguments.

The final formula is obtained by composing the substitutions:
\begin{equation}
\varphi_{\text{final}} = \varphi_0[\theta_1 \circ \theta_2 \circ \theta_3].
\end{equation}

Example~\ref{fig:theta_example} illustrates the resulting stepwise derivation in detail.

\subsection{Entailment tree verification and refinement for recursively verifiable NLI}
Finally, given an entailment tree $\mathcal{T}$, we verify it recursively by turning each subtree into a set of local proof obligations that are checkable by a TP. Each node $n\in\mathcal{T}$ stores a natural language statement $s_n$, its atomic decomposition
$D(n)=\{a_{n1},a_{n2},\dots,a_{n|D(n)|}\}$, and a role indicator
$\rho(n)\in\{\textsc{explanation},\textsc{hypothesis}\}$ that is defined \emph{with respect to the
current subtree under verification}. For a subtree $\mathcal{T}_{\text{sub}}\subseteq\mathcal{T}$ with root $h=\mathrm{root}(\mathcal{T}_{\text{sub}})$ and leaf set
$E=\mathrm{leaves}(\mathcal{T}_{\text{sub}})$, we set
\begin{equation}
\rho(n)=
\begin{cases}
\textsc{explanation} & \text{if } n\in E,\\
\textsc{hypothesis} & \text{if } n=h.
\end{cases}
\end{equation}

\paragraph{From atoms to axioms and lemmas.}
Let $\Phi$ denote autoformalisation from natural language atoms to the target logic. For every explanation node $e\in E$, each atom $a\in D(e)$ is translated into an Isabelle/HOL~\citep{nipkow2002isabelle} axiom (example in Appendix \ref{appendix: implementation_detail}):
\begin{equation}
\text{axiom}(e,a):\ \Phi(a).
\end{equation}
Collecting all explanation atoms yields the axiom set
\begin{equation}
A(E)\ :=\ \bigcup_{e\in E}\ \{\Phi(a)\mid a\in D(e)\}.
\end{equation}
For the hypothesis node $h$, each atom $a\in D(h)$ induces a local proof obligation:
\begin{equation}
\text{lemma}(h,a):\ A(E)\ \vdash\ \Phi(a).
\end{equation}
We accept $h$ as verified for the current subtree if $\mathit{TP}$ certifies
$\text{lemma}(h,a)$ for all atoms $a\in D(h)$.

\paragraph{Recursive verification with localized refinement.}
The verification proceeds bottom-up. Intuitively, once a subtree root is verified, it can be treated
as a support statement for its parent level. Algorithm~\ref{alg:verify_subtree} summarises the
procedure.

When $TP$ cannot discharge a lemma $\text{lemma}(h,a)$ for some atom $a\in D(h)$, we extract
diagnostics from the prover (e.g., the failed goal and relevant constraints) and use them to localise the failure to a small set of implicated explanation nodes $\widehat{E}$. We then prompt an LLM to refine only these implicated nodes, update the subtree accordingly, and re-check the same local obligations. This loop continues until all atoms in $D(h)$ are certified for the subtree root $h$, after which the verified root is promoted and the algorithm proceeds to higher levels. Repeating this process bottom-up yields a recursively verifiable proof certificate for the entire entailment tree.

% =============================================================================
% Explanation Refinement Result Table
% =============================================================================
\setlength{\tabcolsep}{1.5pt}
\begin{table*}[t]
\centering
\small % 
%\resizebox{\textwidth}{!}{%
\begin{tabular}{@{}ll*{5}{cc}@{}}
\toprule
\multirow{2}{*}{\textbf{Dataset}} & \multirow{2}{*}{\textbf{Approach}} &
\multicolumn{2}{c}{\textbf{GPT-4o}} &
\multicolumn{2}{c}{\textbf{GPT-5 nano}} &
\multicolumn{2}{c}{\textbf{Grok-4 fast}} &
\multicolumn{2}{c}{\textbf{Deepseek-V3.1}} &
\multicolumn{2}{c}{\textbf{Qwen3-max}} \\
\cmidrule(lr){3-4} \cmidrule(lr){5-6} \cmidrule(lr){7-8} \cmidrule(lr){9-10} \cmidrule(lr){11-12}
 & & \textit{Init.} & \textit{Fin.}
   & \textit{Init.} & \textit{Fin.}
   & \textit{Init.} & \textit{Fin.}
   & \textit{Init.} & \textit{Fin.}
   & \textit{Init.} & \textit{Fin.} \\
\midrule
% --- FOLIO BLOCK ---
 & Explanation-Refiner & 53.28 & 63.93 & 46.72 & 59.84 & 45.08 & 52.46 & 50.82 & 62.30 & 55.74 & 68.85 \\
 & Faithful-Refiner & 68.03 & 78.69  & 53.28 & 68.03 & 54.92 & 65.57 & 71.31 & 81.15 & 69.67 & 84.43 \\
\rowcolor{gray!15} \cellcolor{white} \multirow{-3}{*}{\textbf{FOLIO}} & \textbf{LLM-TP Tree} & 81.15 & \shortstack{\textbf{90.16}\\\tiny\textcolor{darkgreen}{$\uparrow$26.23}} & 66.39 & \shortstack{\textbf{81.15}\\\tiny\textcolor{darkgreen}{$\uparrow$21.31}} & 62.30 & \shortstack{\textbf{79.51}\\\tiny\textcolor{darkgreen}{$\uparrow$27.05}} & 83.61 & \shortstack{\textbf{92.62}\\\tiny\textcolor{darkgreen}{$\uparrow$30.32}} & 85.25 & \shortstack{\textbf{95.08}\\\tiny\textcolor{darkgreen}{$\uparrow$26.23}} \\
\midrule
% --- ProofWriter BLOCK ---
 & Explanation-Refiner & 62.67 & 76.67 & 59.33 & 73.33 & 53.33 & 69.33 & 52.67 & 66.67 & 64.00 & 78.00 \\
 & Faithful-Refiner & 80.00 & 86.67 & 72.00 & 84.67 & 60.00 & 78.67 & 83.33 & 90.67 & 84.00 & 91.33 \\
\rowcolor{gray!15} \cellcolor{white} \multirow{-3}{*}{\textbf{ProofWriter}} & \textbf{LLM-TP Tree} & 91.33 & \shortstack{\textbf{95.33}\\\tiny\textcolor{darkgreen}{$\uparrow$18.66}} & 85.33 & \shortstack{\textbf{91.33}\\\tiny\textcolor{darkgreen}{$\uparrow$18.00}} & 72.67 & \shortstack{\textbf{90.00}\\\tiny\textcolor{darkgreen}{$\uparrow$20.67}} & 91.33 & \shortstack{\textbf{98.00}\\\tiny\textcolor{darkgreen}{$\uparrow$31.33}} & 92.00 & \shortstack{\textbf{98.00}\\\tiny\textcolor{darkgreen}{$\uparrow$20.00}} \\
\midrule
% --- PrOntoQA BLOCK ---
 & Explanation-Refiner & 64.00 & 78.67 & 60.00 & 74.67 & 61.33 & 73.33 & 62.00 & 74.00 & 64.67 & 78.00 \\
 & Faithful-Refiner & 82.67 & 90.00 & 73.33 & 80.00 & 72.00 & 84.00 & 85.33 & 91.33 & 82.00 & 91.33 \\
\rowcolor{gray!15} \cellcolor{white} \multirow{-3}{*}{\textbf{PrOntoQA}} & \textbf{LLM-TP Tree} & 92.00 & \shortstack{\textbf{97.33}\\\tiny\textcolor{darkgreen}{$\uparrow$18.66}} & 93.33 & \shortstack{\textbf{98.00}\\\tiny\textcolor{darkgreen}{$\uparrow$23.33}} & 86.00 & \shortstack{\textbf{91.33}\\\tiny\textcolor{darkgreen}{$\uparrow$18.00}} & 97.33 & \shortstack{\textbf{100.00}\\\tiny\textcolor{darkgreen}{$\uparrow$26.00}} & 98.00 & \shortstack{\textbf{100.00}\\\tiny\textcolor{darkgreen}{$\uparrow$22.00}} \\
\midrule
% --- EntailmentBank BLOCK ---
 & Explanation-Refiner & 15.33 & 26.67 & 12.00 & 23.33 & 8.67 & 13.33 & 15.33 & 26.67 & 16.67 & 28.00 \\
 & Faithful-Refiner & 23.33 & 44.00 & 17.33 & 47.33 & 14.00 & 46.00 & 25.33 & 48.00 & 22.00 & 52.00 \\
\rowcolor{gray!15} \cellcolor{white} \multirow{-3}{*}{\textbf{EntailmentBank}} & \textbf{LLM-TP Tree} & 21.33 & \shortstack{\textbf{70.00}\\\tiny\textcolor{darkgreen}{$\uparrow$43.33}} & 18.00 & \shortstack{\textbf{74.00}\\\tiny\textcolor{darkgreen}{$\uparrow$50.67}} & 16.67 & \shortstack{\textbf{68.67}\\\tiny\textcolor{darkgreen}{$\uparrow$55.34}} & 22.00 & \shortstack{\textbf{71.33}\\\tiny\textcolor{darkgreen}{$\uparrow$44.66}} & 22.67 & \shortstack{\textbf{78.67}\\\tiny\textcolor{darkgreen}{$\uparrow$50.67}} \\
\bottomrule
\end{tabular}%
%}
\caption{Comparison results on the explanation refinement tasks. \textit{Init.} shows the number of explanation that is initially logically valid. \textit{Fin.} shows the number of explanation that is finally logically valid after refinement. \textbf{Bold} values indicate the best performance. Arrows indicate absolute performance gain of LLM-TP Tree over the baseline.}
\label{tab:explanation_refinement_result}
\end{table*}

\section{Experimental Setup}
We evaluate our neuro-symbolic framework from two complementary perspectives:
(i) explanation refinement apply theorem-prover checking and refine the explanation; and (ii) logical reasoning, which measures standard end-task accuracy to ensure that enhanced verification and refinement do not come at the cost of predictive performance. 

\paragraph{Datasets} We use four datasets ranging from synthetically generated deductive reasoning benchmarks to expert-written real-world corpora: ProofWriter~\citep{tafjord-etal-2021-proofwriter}, PrOntoQA~\citep{PrOntoQA}, FOLIO~\citep{han-etal-2024-folio}, and EntailmentBank~\citep{entailmentbank2021}. Additional dataset details (i.e. number of samples) are provided in Appendix \ref{appendix:dataset_details}.

\paragraph{Baselines and Models}
For explanation refinement, we compare against two state-of-the-art explanation refinement models, Explanation-Refiner~\citep{quan-etal-2024-verification} and Faithful-Refiner~\citep{quan-etal-2025-faithful}.  For logical reasoning, we consider two prompting-based baselines, including a direct standard prompting and chain-of-thought (CoT) prompting. In addition, we compare against two representative neuro-symbolic frameworks, Logic-LM~\citep{pan-etal-2023-logic} and LINC~\citep{olausson-etal-2023-linc}. We evaluate five distinct LLM backbones, spanning a general-purpose chat model (GPT-4o~\citep{openai2024gpt4technicalreport}), models evaluated in non-thinking mode (GPT-5 Nano~\citep{openai2025gpt5}, Grok-4 Fast~\citep{xai2025grok4}), and models evaluated in thinking mode (DeepSeek-V3.1~\citep{deepseekai2025deepseekv3technicalreport}, Qwen3-Max~\citep{yang2025qwen3technicalreport}) and additionally include GPT-3.5 in logical reasoning tasks to match the model coverage reported in prior work. LLM implementation details is described in Appendix \ref{appendix: implementation_detail}.

\section{Empirical Results and Evaluation}

% =============================================================================
\begin{table}[t]
\centering
\small
\setlength{\tabcolsep}{5pt}
\renewcommand{\arraystretch}{1.05}
\begin{tabular}{lcccc}
\toprule
\textbf{Dataset} & \textbf{ER} & \textbf{FR} & \textbf{LT} & \textbf{LT w/o AD} \\
\midrule
FOLIO          & 4.27 & 3.75 & 3.04 & 3.57 \\
ProofWriter    & 3.47 & 3.08 & 2.42 & 2.88 \\
PrOntoQA       & 3.36 & 2.73 & 2.04 & 2.44 \\
EntailmentBank & 4.97 & 4.46 & 4.04 & 4.35 \\
\midrule
\textbf{Avg.}  & 4.02 & 3.51 & 2.88 & 3.31 \\
\bottomrule
\end{tabular}
\caption{Average refinement iterations required averaged over five LLM backbones. Comparison between ER (Explanation-Refiner), FR (Faithful-Refiner), LT (LLM-TP Tree) and LT w/o AD (LT without atomic decomposition)}
\label{tab:avg_iterations_by_dataset}
\end{table}

\paragraph{LLM-TP Tree effectively verifies and refines explanations for NLI.}
Table~\ref{tab:explanation_refinement_result} reports the main results on the explanation refinement task. Across all datasets and all LLM backbones, LLM-TP Tree achieves the best final refinement performance (\textit{Fin.}), indicating that it is consistently more effective at producing a checkable witness for explanation-based NLI. Averaged over backbone LLMs, LLM-TP Tree improves the final verified rate by 26.23\%, 21.73\%, 21.60\%, and 48.93\% on each dataset compared to Explanation-Refiner. The effect is most pronounced on EntailmentBank: while holistic refiners remain below 52\% final validity, LLM-TP Tree reaches 68.67--78.67\% across backbones. Beyond final refinement quality, the \textit{Init.} column probes whether a backbone can produce an initially provable explanatory chain before any refinement cycle. On synthetic benchmarks such as
ProofWriter and PrOntoQA, LLM-TP Tree yields substantially higher initial validity than holistic refiners; for instance, on ProofWriter with Qwen3-max, LLM-TP Tree attains 92.00\% \textit{Init.} versus 64.00\% for Explanation-Refiner. Notably, EntailmentBank exhibits low initial validity across methods, suggesting that constructing globally coherent verifiable chains is difficult under more realistic multi-hop explanations. In this regime, our proposed mechanism yields the largest gains after refinement.

\begin{figure*}[t]
    \centering
    \begin{subfigure}[b]{0.24\textwidth}
        \centering
        \includegraphics[width=\textwidth]{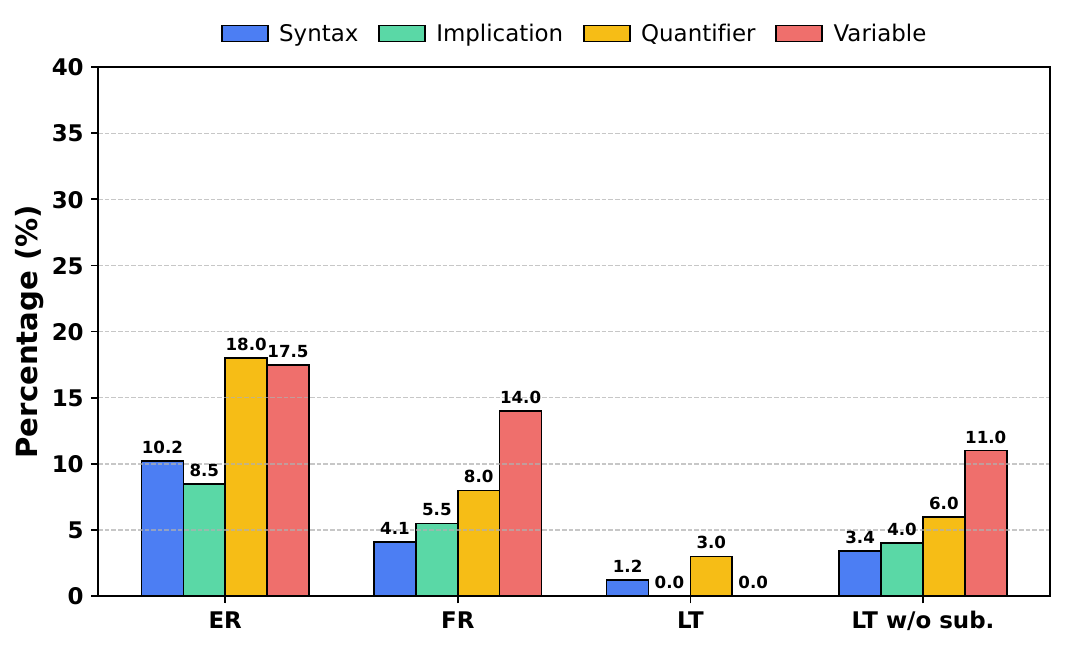}
        \caption{FOLIO}
        \label{fig:folio_syntax_error}
    \end{subfigure}
    \hfill
    \begin{subfigure}[b]{0.24\textwidth}
        \centering
        \includegraphics[width=\textwidth]{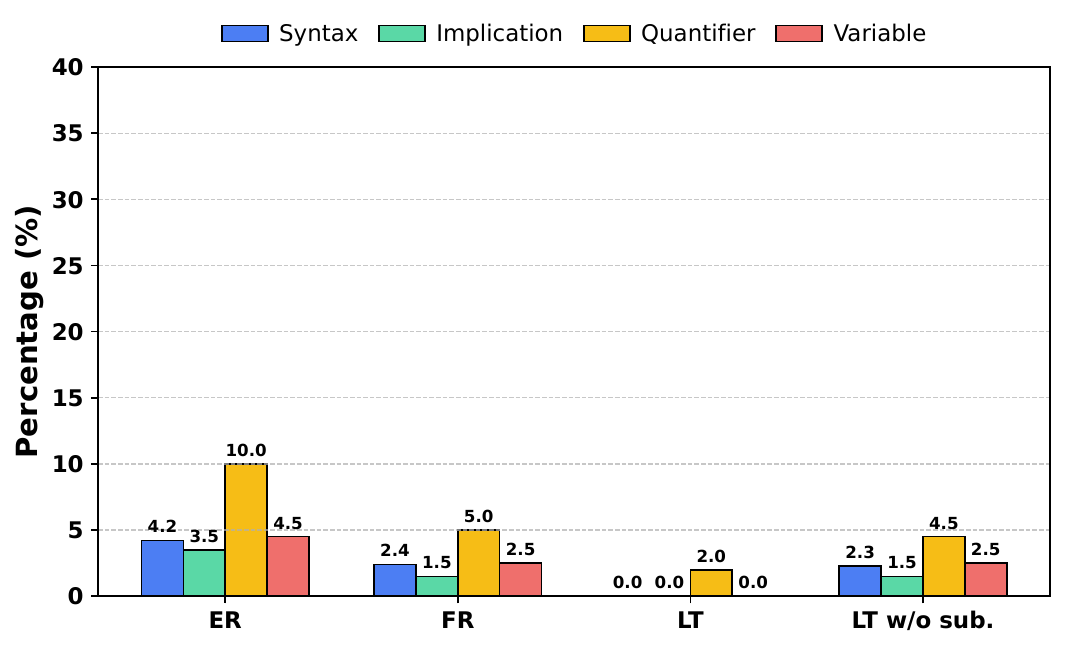}
        \caption{ProofWriter}
        \label{fig:proofwriter_syntax}
    \end{subfigure}
    \hfill
    \begin{subfigure}[b]{0.24\textwidth}
        \centering
        \includegraphics[width=\textwidth]{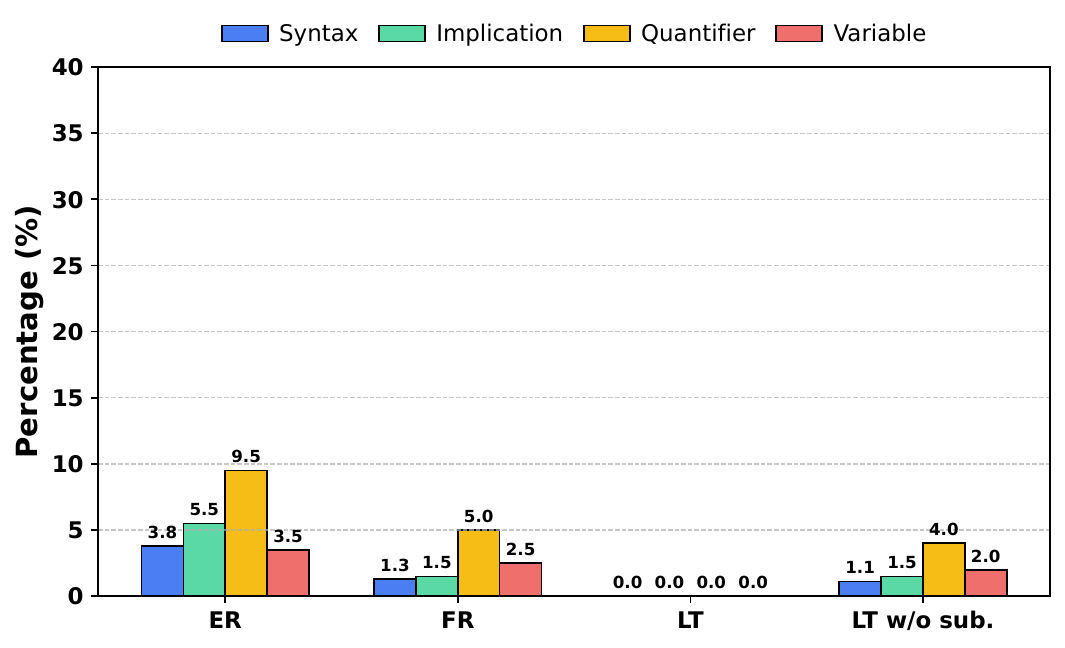}
        \caption{PrOntoQA}
        \label{fig:prontoqa_syntax}
    \end{subfigure}
    \hfill
    \begin{subfigure}[b]{0.24\textwidth}
        \centering
        \includegraphics[width=\textwidth]{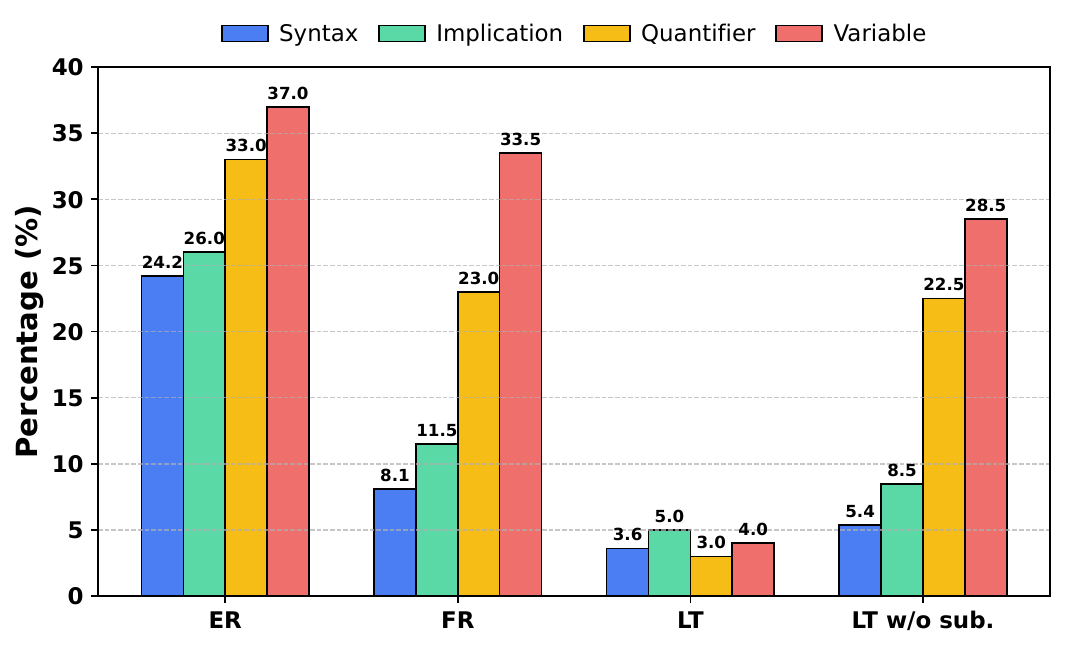}
        \caption{EntailmentBank}
        \label{fig:entailmentbank_syntax}
    \end{subfigure}
    \caption{Distribution of the autoformlisation errors (syntax, implication, quantifier, variable in Isabelle/HOL Theories.}
    \label{fig:comparison_syntax_error}
\end{figure*}

% 0.71, 0.66, 0.694, 0.422

\paragraph{Atomic decomposition improves the efficiency of verification-driven explanation refinement.}  We report the average number of refinement iterations, averaged across different LLM backbones, required to refine an explanation for each dataset, as shown in Table~\ref{tab:avg_iterations_by_dataset}. Across all four datasets (FOLIO, ProofWriter, PrOntoQA, and EntailmentBank), LT converges in fewer iterations than baseline refiners, reducing the required iterations by an average of 1.14 relative to ER and by an average of 0.63 relative to Faithful-Refiner (FR). To isolate the advantages of atomic decomposition, we run an ablation that disables atomic sentence splitting. Relative to the full LT system, removing atomic decomposition increases the number of required refinement iterations by between 0.31 and 0.53 . This ablation indicates that atomic decomposition contributes materially to LT's efficiency gains: by splitting premises and intermediate conclusions into atomic subclaims, verification failures can be localised to a small set of fine-grained hypotheses, enabling targeted repairs of the minimal logical mismatch and thereby reducing both the number of refinement iterations and the cost of each verification/refinement pass. We also report the mean running time per refinement iteration (including theorem-proving time) under different LLM backbones, and provide the full per-backbone iteration results in Appendix~\ref{appendix: inference_time_thinking}.

\begin{table}[t]
\centering
\small
\setlength{\tabcolsep}{5pt}
\renewcommand{\arraystretch}{1.05}
\begin{tabular}{lcccc}
\toprule
\textbf{Dataset} & \textbf{ER} & \textbf{FR} & \textbf{LT} & \textbf{LT w/o sub.} \\
\midrule
FOLIO          & 0.766 & 0.822 & 0.856 & 0.828 \\
ProofWriter    & 0.834 & 0.892 & 0.940 & 0.914 \\
PrOntoQA       & 0.824 & 0.918 & 0.952 & 0.924 \\
EntailmentBank & 0.614 & 0.726 & 0.824 & 0.770 \\
\midrule
\textbf{Avg.}  & 0.760 & 0.840 & 0.893 & 0.859 \\
\bottomrule
\end{tabular}
\caption{Average autoformalisation faithfulness (cosine similarity) averaged over five LLM backbones. Comparison between ER (Explanation-Refiner), FR (Faithful-Refiner), LT (LLM-TP Tree) and LT w/o sub. (LT without $\theta$-substitution autoformalisation).}
\label{tab:avg_faithfulness_by_dataset}
\end{table}

\paragraph{Multi-step $\theta$-substitution autoformalisation improves semantic faithfulness.} A faithful autoformalisation should preserve the semantic content of the original sentence as much as possible. Following the rule-based informalisation procedure by \citet{quan-etal-2025-faithful}, we quantify faithfulness by converting each predicted logical form back into natural language and computing the cosine similarity between the informalised sentence and the original sentence. Table~\ref{tab:avg_faithfulness_by_dataset} summarises the results. Our approach achieves the highest average faithfulness across all datasets, outperforming Explanation-Refiner by an average of 0.13 and Faithful-Refiner by an average of 0.053. As expected, faithfulness is generally higher on synthetic benchmarks, whose language more closely follows rule-based templates, whereas EntailmentBank is more challenging due to its more naturalistic multi-hop explanations. Notably, our method yields the largest margin on EntailmentBank, indicating stronger robustness of $\theta$-substitution autoformalisation under distributional and linguistic variability. We further perform an ablation by removing $\theta$-substitution (LT w/o sub.). This variant shows a consistent drop in similarity of an average of 0.034 across the same datasets compared to the full model. We also shows the full ablation study results in Appendix \ref{appendix: ablation}

\subsection{Failure and Error Analysis}
\begin{figure}[!ht]
    \centering
    \begin{subfigure}[t]{0.23\textwidth} 
        \centering
        \includegraphics[width=\textwidth]{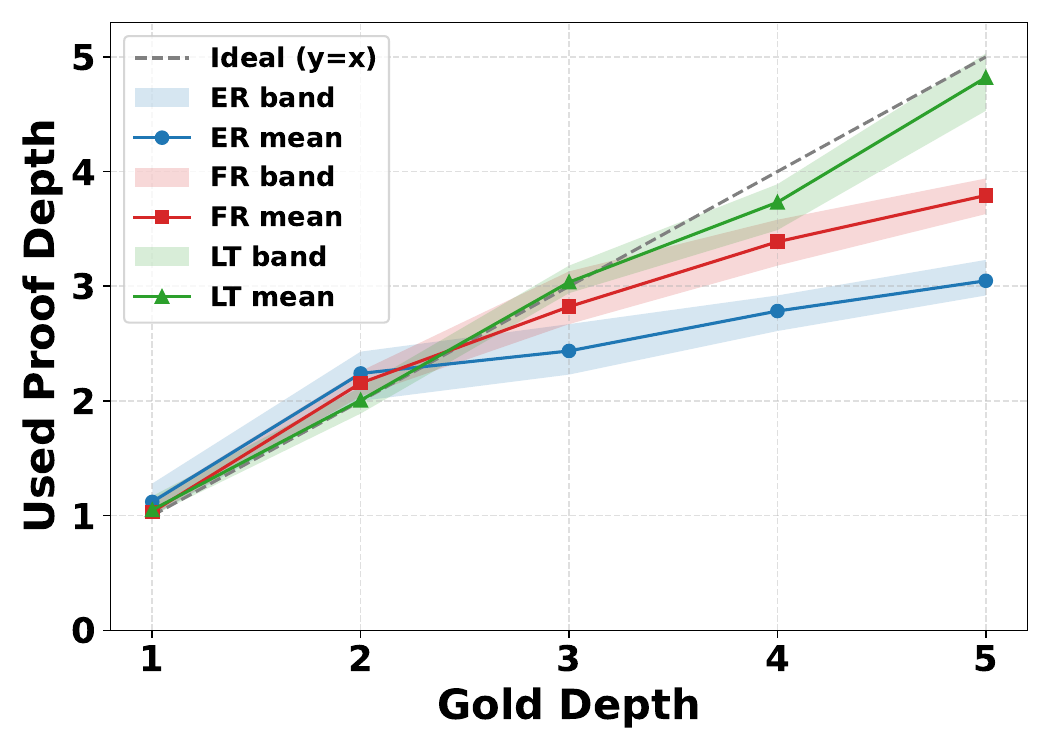}
        \caption{ProofWriter - Refined}
        \label{fig:proofwriter_proof_depth_refined}
    \end{subfigure}
    \hfill  
    \begin{subfigure}[t]{0.23\textwidth} 
        \centering
        \includegraphics[width=\textwidth]{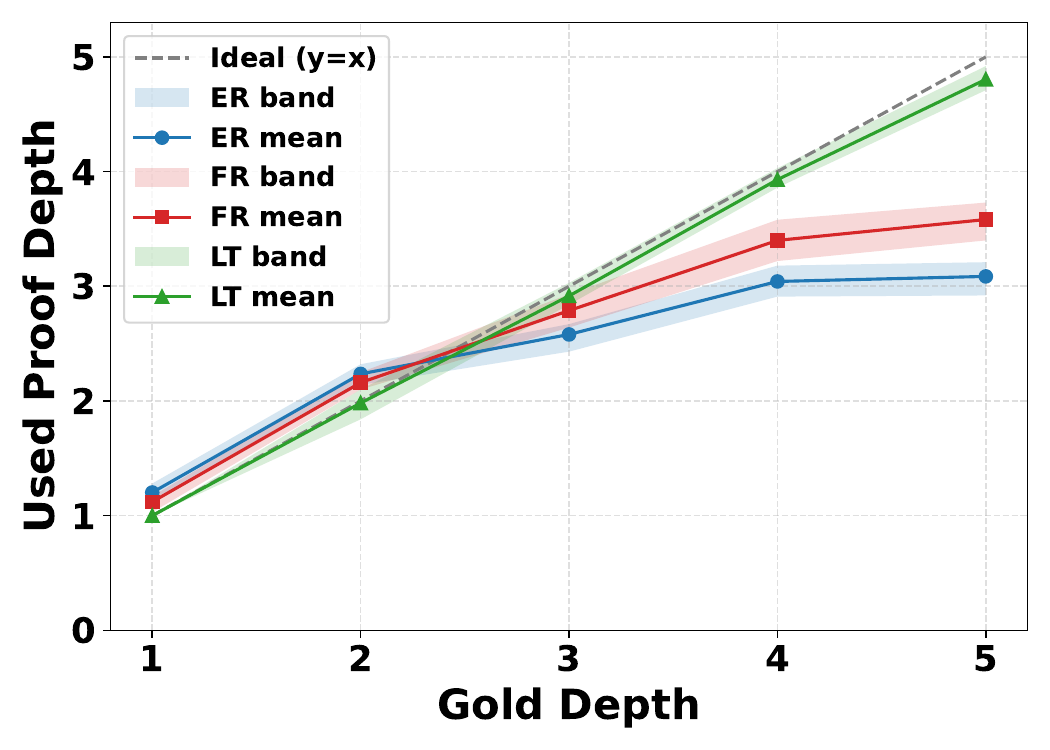}
        \caption{EntailmentBank - Refined}  
        \label{fig:entailmentbank_proof_depth_refined}
    \end{subfigure}
    \hfill  
    \begin{subfigure}[t]{0.23\textwidth} 
        \centering
        \includegraphics[width=\textwidth]{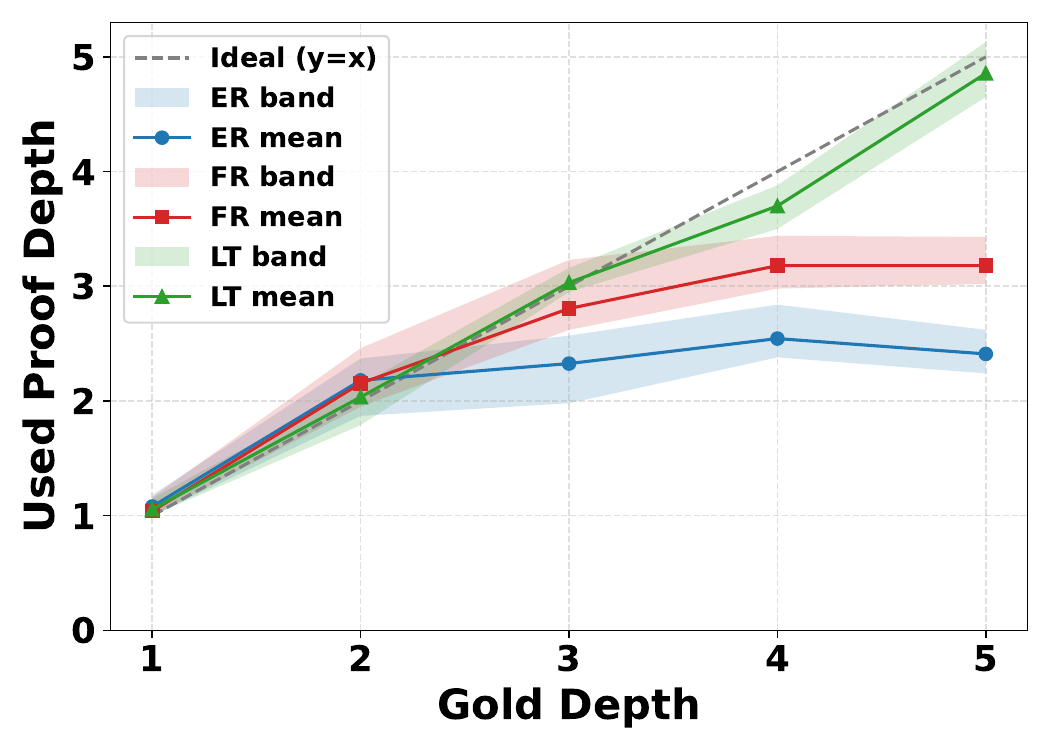}
        \caption{ProofWriter - Unrefined}
        \label{fig:proofwriter_proof_depth_unrefined}
    \end{subfigure}
    \hfill  
    \begin{subfigure}[t]{0.23\textwidth} 
        \centering
        \includegraphics[width=\textwidth]{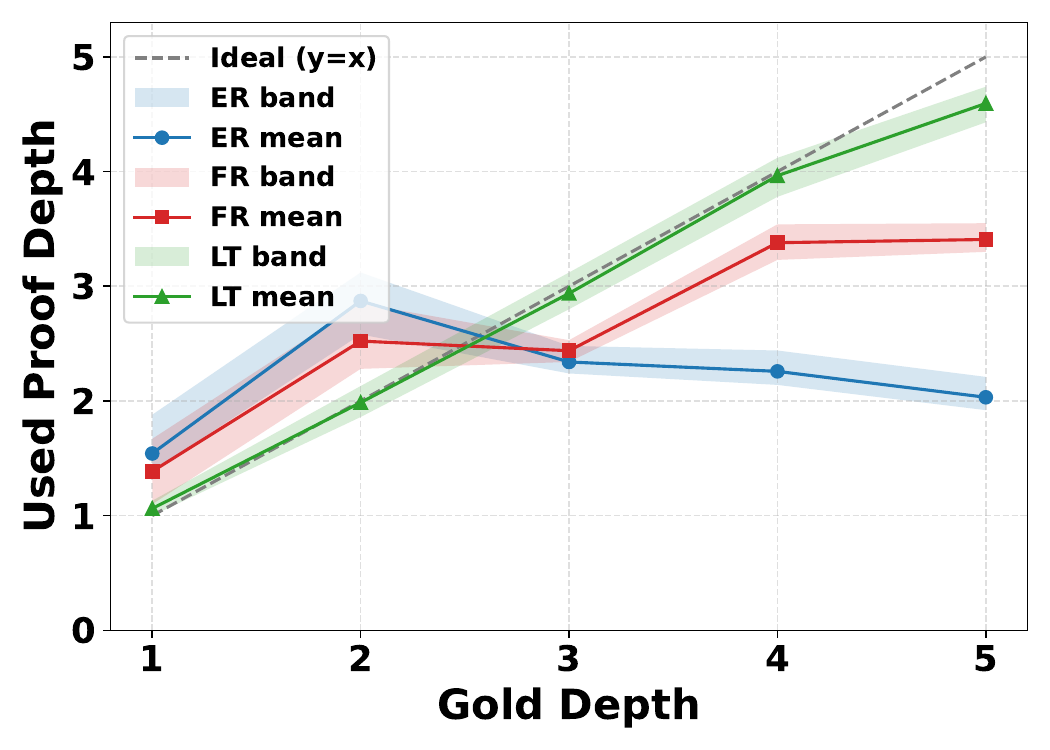}
        \caption{EntailmentBank - Unrefined}  
        \label{fig:entailmentbank_proof_depth_unrefined}
    \end{subfigure}
    \caption{Proof depth alignment between gold and actual constructed proof depths across frameworks. The solid curve shows the mean used depth averaged over five LLMs, while the shaded band shows the range across backbones. Top: Trends in refined cases. Bottom: Trends in unrefined cases.}
    \label{fig:proof_depth}
\end{figure}

\begin{figure*}[!ht]
    \centering
    \begin{subfigure}[b]{0.31\textwidth}
        \centering
        \includegraphics[width=\textwidth]{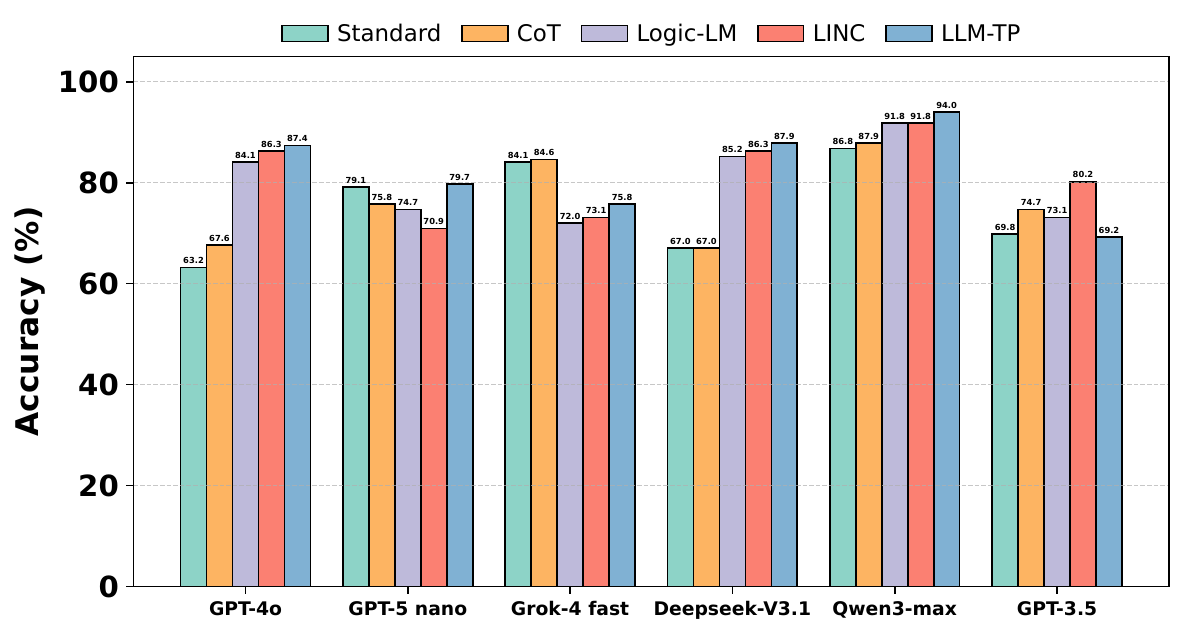}
        \caption{FOLIO}
        \label{fig:folio_logical_accuracy}
    \end{subfigure}
    \hfill
    \begin{subfigure}[b]{0.31\textwidth}
        \centering
        \includegraphics[width=\textwidth]{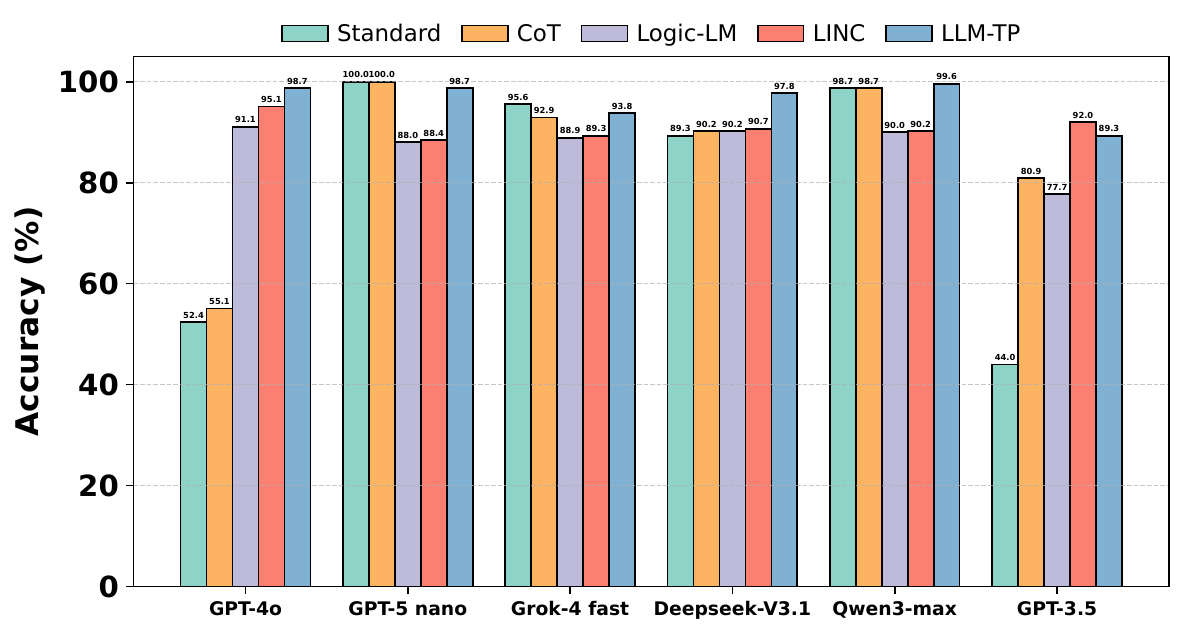}
        \caption{ProofWriter}
        \label{fig:proofwriter_logical_accuracy}
    \end{subfigure}
    \hfill
    \begin{subfigure}[b]{0.31\textwidth}
        \centering
        \includegraphics[width=\textwidth]{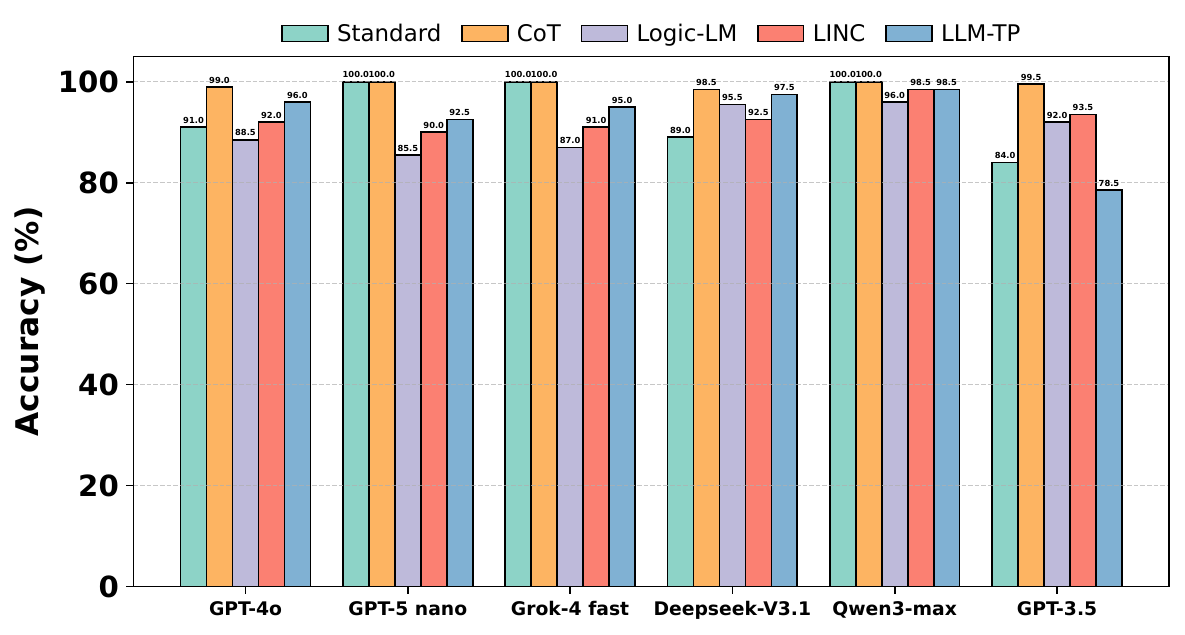}
        \caption{PrOntoQA}
        \label{fig:prontoqa_logical_accuracy}
    \end{subfigure}
    \caption{Comparison on the logical reasoning accuracy tasks.}
    \label{fig:logical_reasoning}
\end{figure*}

We analyse refinement drift using proof-depth alignment. Figure ~\ref{fig:proof_depth} measure proof-depth alignment by comparing the TP-extracted used proof depth of each explanation with the dataset-provided gold depth (ideal $y{=}x$). In refined cases, all frameworks show a positive trend that is close to the ideal $y{=}x$ line. By contrast, in unrefined cases, ER and FR exhibit a clear refinement drift in the form of depth compression: the used depth grows slowly, saturates, or even decreases, increasingly undershooting the gold depth. This indicates that holistic refiners tend to collapse longer multi-step explanations into shorter proofs. In contrast, LLM-TP Tree tracks the gold depth much more closely suggesting that its refinement process better preserves proof-step granularity in longer-horizon inference. Detailed distributions are provided in Appendix~\ref{appendix:proof_depth}. We further analyse autoformalisation failures by categorising errors in the generated Isabelle/HOL theories into four types (details in Appendix \ref{appendix:error_type}) across 200 sampled theories. Figure~\ref{fig:comparison_syntax_error} shows that LLM-TP Tree consistently yields the lowest error rates across datasets, with the largest margin on the more naturalistic EntailmentBank. Removing $\theta$-substitution (LT w/o sub.) leads to a clear increase in semantic and structural errors, particularly quantifier and variable errors on EntailmentBank, highlighting the importance of staged substitution autoformalisation for maintaining a prover-compatible logical structure.

\subsection{Logical Reasoning}

Figure~\ref{fig:logical_reasoning} reports accuracy on three logical reasoning benchmarks. Across all datasets and LLM backbones, LLM-TP Tree is the strongest neuro-symbolic method: compared with Logic-LM and LINC, it achieves higher overall accuracy, with the largest gains on the more challenging settings where multi-step composition is essential. These results suggest that explicit theorem proving, coupled with a tree-structured verification procedure, provides a reliable supervision signal for reducing spurious inferences in existing neuro-symbolic pipelines, especially when the backbone model can produce sufficiently accurate autoformalisation.

At the same time, prompting-only baselines such as CoT can still be competitive and may outperform neuro-symbolic approaches on certain datasets. In particular, on PrOntoQA, Standard and CoT prompting
remain consistently strong, leaving limited headroom for additional symbolic verification to improve raw accuracy. This pattern is consistent with two factors: some benchmarks are close to saturation under strong prompting, and neuro-symbolic methods introduce an additional failure mode through autoformalisation, where an incomplete or incorrect theory prevents the prover from establishing a label. Importantly, even in regimes where CoT is stronger in accuracy, LLM-TP Tree provides an advantage that prompting-only methods do not: when successful, it produces solver-checkable proof certificates and intermediate structure that support auditability and diagnosis, which are central to verifiable NLI.

\section{Related Work}
Neuro-symbolic pipelines have shown strong effectiveness on multi-hop reasoning, as they combine learned language understanding with symbolic reasoning to support complex, multi-step inference. Early work such as \citet{weber-etal-2019-nlprolog} integrates a symbolic reasoner (Prolog) with a rule-learning component to solve multi-hop reasoning problems. More recently, researchers have increasingly combined LLMs with symbolic solvers across a range of logical reasoning tasks. For example, \citet{pan-etal-2023-logic} and \citet{olausson-etal-2023-linc} autoformalise natural-language contexts into symbolic representations and then apply external solvers to perform deductive reasoning over the resulting formal structures. Along a related direction, \citet{xu-etal-2025-aristotle} proposes a decompose-search-resolve framework that breaks a problem into smaller logical substructures and performs structured search to resolve each reasoning step. Prover feedback has also been used to improve reliability: \citet{quan-etal-2025-peirce} extracts
erroneous proof steps from solver feedback and iteratively refines reasoning in a loop. In contrast to endpoint-focused verification, our work targets recursively verifiable NLI by enforcing chain-level checking and localised refinement over structured entailment trees.

\section{Conclusion}
We presented {LLM-TP Tree, a neuro-symbolic framework for recursively verifiable NLI that follows a decompose-and-formalise paradigm. By combining entailment-tree verification with entailment-preserving atomic decomposition and $\theta$-substitution autoformalisation, our method enables localised, refinement that targets the implicated part of an explanation rather than regenerating it holistically. Experiments on four datasets with multiple LLM backbones show improved TP-verified explanation rates, more faithful autoformalisation, and reduced refinement drift while maintaining strong end-task reasoning accuracy. In future work, we will extend this framework to broader naturalistic settings, improve the robustness of entailment-tree construction and diagnostic localisation, and further reduce verification cost through better reuse of verified subtrees.

\section*{Limitations}
Our framework produces solver-checkable certificates, but the guarantee is necessarily conditional on the chosen autoformalisation function $\Phi$, the target formalism, and the capabilities of the underlying theorem prover. In particular, a verification can be valid in the induced formal theory while still being partially misaligned with the original natural language intent if $\Phi$ introduces subtle scope, polarity, or role mismatches that are not exposed by our current faithfulness checks. This dependence is amplified in naturalistic NLI, where missing background knowledge, implicit commonsense assumptions, or discourse phenomena (e.g., coreference, modality, temporality) may not be representable in the event-based logical form, making some instances difficult to certify without additional axioms or richer semantics.  The effectiveness of localised refinement also depends on the quality of the intermediate structures
proposed by the LLM and on the ability to map prover diagnostics back to the responsible region. When the initial entailment tree omits critical intermediate conclusions or introduces a misleading proof plan, bottom-up checking can localise which obligations are not discharged but may not reliably recover the globally correct chain without substantial restructuring.  Finally, although our approach reduces expensive global regeneration, it still incurs non-trivial
compute due to repeated LLM calls, repeated autoformalisation, and repeated prover invocations. Prover timeouts or inconclusive outcomes can remain a practical bottleneck on harder instances, and our current system does not provide a formal convergence guarantee to the minimal or unique refined explanation.

\bibliography{custom}

\appendix

\section{Dataset Details}
\label{appendix:dataset_details}

\subsection{Explanation Refinement}

\paragraph{ProofWriter.}
ProofWriter~\citep{tafjord-etal-2021-proofwriter} is a synthetically generated deductive reasoning benchmark where each instance contains a natural language theory (facts and rules), a query, and a gold truth label, with controllable multi-hop proof depth. For explanation refinement, we sample 150 instances and stratify the sampling to cover a range of gold proof depths $d\in\{1,2,3,4,5\}$ on the ProofWriter D5 split under the open-world assumption. To avoid ambiguity in verification-oriented refinement, we restrict to boolean-labelled instances and exclude indeterminate cases (e.g., \texttt{Unknown} under open-world settings). Each instance is cast into an NLI-style pair $(P,h)$, where $P$ is the provided set of facts/rules and $h$ is the hypothesis statement.

\paragraph{PrOntoQA.}
PrOntoQA~\citep{PrOntoQA} is a synthetic dataset designed to evaluate deductive reasoning in controlled settings. Following the Logic-LM~\citep{pan-etal-2023-logic} setup, we use the most challenging five-hop subset (fictional-entity reasoning). For explanation refinement, we randomly sample 150 instances from this subset. Since this subset is binary in nature, we keep boolean-labelled instances and cast each instance into the NLI-style format $(P,h)$.

\paragraph{FOLIO.}
FOLIO~\citep{han-etal-2024-folio} is a human-curated first-order logic reasoning benchmark, pairing natural-language statements with formal logical structure to support theorem-prover-based verification. For explanation refinement, we use the evaluation split commonly adopted by prior neuro-symbolic work after preprocessing, and remove indeterminate labels to keep a boolean setting. Concretely, we start from 182 evaluation instances after filtering out 22 erroneous examples identified by \citet{olausson-etal-2023-linc} and exclude those with uncertain labels, yielding 122 instances. We then treat the remaining instances as NLI pairs $(P,h)$ for verification and refinement.

\paragraph{EntailmentBank.}
EntailmentBank~\citep{entailmentbank2021} is a multi-hop entailment benchmark that provides naturalistic premises and hypotheses together with structured explanation annotations (intermediate conclusions and entailment structure). For explanation refinement, we sample 150 instances and select examples spanning different gold proof depths to ensure coverage of both shallow and longer-horizon multi-step chains. Each instance is cast into $(P,h)$, where $P$ is the set of supporting statements (facts) and $h$ is the hypothesis.

\subsection{Logical Reasoning}

\paragraph{ProofWriter.}
For logical reasoning, we evaluate on the ProofWriter D5 split under the open-world assumption to retain the three-way label space (\texttt{True/False/Unknown}). We sample 225 instances stratified by gold reasoning depth $d\in\{1,2,3,4,5\}$, with 45 instances per depth. Within each depth bucket, we balance labels by sampling 15 instances per class.

\paragraph{FOLIO.}
For logical reasoning, the original validation split contains 204 instances, while we select the 182 evaluation instances after filtering out 22 erroneous examples identified by \citet{olausson-etal-2023-linc} for evaluation. We evaluate on this 182-instance set under the dataset’s original label space.

\paragraph{PrOntoQA.}
For logical reasoning, we follow prior work~\citep{pan-etal-2023-logic} and evaluate on the five-hop PrOntoQA subset. We randomly sample 200 instances from this subset (binary labels), which represents the most challenging multi-hop configuration in the benchmark.

\section{Autoformalisation Errors}
\label{appendix:error_type}
Following the error taxonomy in \citet{quan-etal-2025-faithful}, we categorise autoformalisation errors into four broad types: \textit{syntax}, \textit{implication}, \textit{quantifier}, and \textit{variable} errors. 
We define an autoformalisation error as any mismatch between the intended logical meaning of the input sentence(s) (as captured by a reference logic template or gold formalisation) and the produced formal statement. 
Notably, these categories differ in how easily they can be detected: syntax errors are often surfaced immediately by the theorem prover, whereas the remaining three categories typically produce well-typed, well-formed formulas that may still be semantically unfaithful and therefore require semantic-level checking. 

\paragraph{Syntax errors.} Syntax errors are those that prevent the proof assistant from parsing or type-checking the generated statement, and are therefore usually the easiest to detect.  Typical causes include missing brackets, malformed binder syntax, incorrect use of infix operators, and type unification conflicts (e.g., using an event-typed variable where an entity is expected).  Such errors are often directly indicated by the theorem prover’s feedback (parse errors, unknown constants, failed unification, or unsatisfied type-class constraints), making them amenable to automatic filtering.

\paragraph{Implication errors.} Implication errors occur when the generated formula has a valid syntax and type, but its \emph{logical structure} encodes the wrong entailment relation compared to the intended template. 
This includes incorrect direction of implication, incorrect connective choice (e.g., $\wedge$ vs.\ $\rightarrow$), or wrong nesting/precedence that changes the meaning. 
These errors are difficult for theorem provers to flag because the statement may still be perfectly consistent and provable under some background axioms, while representing a different claim.

\paragraph{Quantifier errors.}
Quantifier errors arise when the autoformalisation introduces incorrect quantifiers ($\forall$ vs.\ $\exists$), incorrect quantifier scope, or incorrect ordering of quantifiers. 
These errors are especially harmful because they often produce statements that are strictly stronger or weaker than intended while remaining syntactically valid. Moreover, theorem provers typically do not report quantifier mismatches as errors; the formula is well-formed, yet its truth conditions can differ dramatically.

\paragraph{Variable errors.}
Variable errors occur when logical variables are incorrectly assigned to predicate arguments, resulting in role confusion or broken discourse links. 
This category covers a wide range of semantic binding mistakes, including swapping semantic roles (e.g., agent vs.\ patient), incorrectly linking pronouns/coreference, using the wrong event/entity variable in a clause, or mistakenly reusing a variable that should be distinct.  Variable errors may be subtle because they can remain type-correct (e.g., swapping two entity variables still type-checks) yet invert the intended meaning.

\section{Algorithm}

\begin{algorithm}[t]
\small
\caption{VerifySubtree$(\mathcal{T}_{\text{sub}}, \mathit{TP})$}
\label{alg:verify_subtree}
\begin{algorithmic}[1]
\STATE $h \leftarrow \mathrm{root}(\mathcal{T}_{\text{sub}})$ \COMMENT{current hypothesis node}
\STATE $E \leftarrow \mathrm{leaves}(\mathcal{T}_{\text{sub}})$ \COMMENT{current explanation nodes}
\STATE Construct $A(E)=\bigcup_{e\in E}\{\Phi(a)\mid a\in D(e)\}$ \COMMENT{axioms}
\FOR{each atom $a\in D(h)$}
    \STATE Define the goal $\tau \leftarrow \Phi(a)$
    \IF{$\mathit{TP}.\mathrm{prove}\big(A(E)\vdash \tau\big)=\textsc{fail}$}
        \STATE $\mathrm{diag}\leftarrow \mathit{TP}.\mathrm{diagnostics}()$
        \STATE Identify implicated explanation nodes $\widehat{E}\subseteq E$ from $\mathrm{diag}$
        \STATE Refine $\widehat{E}$ using an LLM conditioned on $(\widehat{E},\mathrm{diag})$
        \STATE Update $\mathcal{T}_{\text{sub}}$; recompute $D(\cdot)$ and re-autoformalise $\Phi(\cdot)$
        \STATE \textbf{return} VerifySubtree$(\mathcal{T}_{\text{sub}}, \mathit{TP})$ \COMMENT{retry locally}
    \ENDIF
\ENDFOR
\STATE Mark $h$ as verified
\STATE \textbf{return} $h$ \COMMENT{promote as support for parent level}
\end{algorithmic}
\end{algorithm}

\section{Isabelle/HOL and LLMs Implementation Detail}
\label{appendix: implementation_detail}
We use the \texttt{isabelle-client} Python package~\citep{shminke2022pythonclientisabelleserver} to run Isabelle/HOL as a real-time server and to retrieve prover responses. We access all LLM backbones via API calls with the following model versions: GPT-4o (\texttt{gpt-4o}), GPT-5 nano (\texttt{gpt-5-nano}), Grok-4 fast (\texttt{grok-4-fast}), Deepseek-V3.1
(\texttt{DeepSeek-v3.1-terminus}), and Qwen3-Max (\texttt{qwen3-max-preview}). For all non-thinking models, we set the temperature to 0. We also do not limit any thinking effort for Deepseek-V3.1 and Qwen3-Max.

We build the Isabelle/HOL theory following the approach in~\citep{quan-etal-2025-faithful}. An example of the constructed Isabelle/HOL theory is shown in Figure~\ref{isabelle_example}. We then apply atomic decomposition to each axiom and to the hypothesis. The decomposed hypothesis sentence and the decomposed axioms are then forming a new theory, which we use to prove theorems from each decomposed hypothesis atoms.

\begin{figure*}
\begin{lstlisting}
theory data_0
imports Main

begin

typedecl entity
typedecl event

consts
  Melting :: "event $\Rightarrow$ bool"
  Change :: "event $\Rightarrow$ bool"
  Source :: "event $\Rightarrow$ entity $\Rightarrow$ bool"
  Destination :: "event $\Rightarrow$ entity $\Rightarrow$ bool"
  Solid :: "entity $\Rightarrow$ bool"
  Liquid :: "entity $\Rightarrow$ bool"
  IncreaseHeatEnergy :: "entity $\Rightarrow$ bool"
  By :: "event $\Rightarrow$ entity $\Rightarrow$ bool"
  Chocolate :: "entity $\Rightarrow$ bool"
  Melts :: "event $\Rightarrow$ bool"
  Agent :: "event $\Rightarrow$ entity $\Rightarrow$ bool"
  In :: "event $\Rightarrow$ entity $\Rightarrow$ bool"
  Sunlight :: "entity $\Rightarrow$ bool"

(* Explanation 1: melting means changing from a solid to a liquid by increasing heat energy *)
axiomatization where
  explanation_1: "$\forall$ e x y z. Melting e $\leftrightarrow$ (Change e $\wedge$ Source e x $\wedge$ Destination e y $\wedge$ Solid x $\wedge$ Liquid y $\wedge$ IncreaseHeatEnergy z $\wedge$ By e z)"

(* Explanation 2: chocolate melts in the sunlight *)
axiomatization where
  explanation_2: "$\exists$ x e. Chocolate x $\wedge$ Melts e $\wedge$ Agent e x $\wedge$ In e Sunlight"

theorem hypothesis:
  assumes asm: "Chocolate x $\wedge$ Solid y $\wedge$ Liquid z"
  (* Hypothesis: chocolate changes from a solid to a liquid in the sunlight *)
  shows False
  sledgehammer
  oops

end
\end{lstlisting}
\caption{An example of the constructed Isabelle/HOL Theory.} 
\label{isabelle_example}
\end{figure*}

\section{Ablation Study on Explanation Refinement Task}
\label{appendix: ablation}
\setlength{\tabcolsep}{2pt}
\begin{table}[t]
\centering
\scriptsize
\renewcommand{\arraystretch}{0.92}

\begin{tabular}{@{}lcccccc@{}}
\toprule
\multirow{2}{*}{\textbf{Dataset}} &
\multicolumn{2}{c}{\textbf{LT}} &
\multicolumn{2}{c}{\textbf{w/o Atom}} &
\multicolumn{2}{c}{\textbf{w/o $\theta$}} \\
\cmidrule(lr){2-3}\cmidrule(lr){4-5}\cmidrule(lr){6-7}
& \textit{Init.} & \textit{Fin.} & \textit{Init.} & \textit{Fin.} & \textit{Init.} & \textit{Fin.} \\
\midrule

\multicolumn{7}{@{}l}{\textbf{Backbone: GPT-4o}}\\
FOLIO
& 81.15 & 90.16
& \dec{71.31}{9.84} & \dec{86.07}{4.09}
& \dec{75.41}{5.74} & \dec{88.52}{1.64} \\
ProofWriter
& 91.33 & 95.33
& \dec{87.33}{4.00} & \dec{92.00}{3.33}
& \dec{89.33}{2.00} & \dec{94.67}{0.66} \\
PrOntoQA
& 92.00 & 97.33
& \dec{86.67}{5.33} & \dec{92.67}{4.66}
& \dec{90.00}{2.00} & \dec{94.00}{3.33} \\
EntailmentBank
& 21.33 & 70.00
& \dec{17.33}{4.00} & \dec{53.33}{16.67}
& \dec{19.33}{2.00} & \dec{61.33}{8.67} \\
\midrule

\multicolumn{7}{@{}l}{\textbf{Backbone: Qwen3-max}}\\
FOLIO
& 85.25 & 95.08
& \dec{72.13}{13.12} & \dec{86.89}{8.19}
& \dec{79.51}{5.74}  & \dec{90.98}{4.10} \\
ProofWriter
& 92.00 & 98.00
& \dec{88.00}{4.00} & \dec{93.33}{4.67}
& \dec{88.00}{4.00} & \dec{96.00}{2.00} \\
PrOntoQA
& 98.00 & 100.00
& \dec{86.67}{11.33} & \dec{94.00}{6.00}
& \dec{92.00}{6.00}  & \dec{96.00}{4.00} \\
EntailmentBank
& 22.67 & 78.67
& \dec{18.00}{4.67} & \dec{61.33}{17.34}
& \dec{20.00}{2.67} & \dec{71.33}{7.34} \\
\bottomrule
\end{tabular}

\vspace{-2mm}
\caption{\small Ablations of LLM-TP Tree on GPT-4o and Qwen3-max (Init/Fin verified rate, \%).
Red arrows indicate the absolute decrease compared to LT under the same backbone.}
\label{tab:ablation_two_backbones}
\vspace{-3mm}
\end{table}

Table~\ref{tab:ablation_two_backbones} reports ablations of LLM-TP Tree (LT) on GPT-4o and Qwen3-max, using \textit{Init.}/\textit{Fin.} logically valid explanation rates. Removing either of the two proposed components, atomic decomposition or multi-step $\theta$-substitution autoformalisation, consistently degrades across all datasets. Across datasets, \textit{Init.} decreases by 4.00 to 13.12 points and \textit{Fin.} decreases by 3.33 to 17.34 points, with the most pronounced degradation on EntailmentBank. Removing $\theta$-substitution also leads to a consistent drop, although smaller than w/o Atom indicating that multi-step substitution contributes materially to producing prover-compatible formalisations.

\section{Full results: Model efficiency Across LLM Backbones}
\label{appendix: inference_time_thinking}
Figure \ref{fig:folio_iteration}, \ref{fig:proofwriter_iteration} and \ref{fig:prontoqa_iteration} shows average number of iterations required to refine an explanation across all LLMs. Figure \ref{fig:folio_inference_time}, \ref{fig:proofwriter_inference_time}, \ref{fig:prontoqa_inference_time} and \ref{fig:comparison_efficiency_thinking} shows average inference time per iteration in both thinking and non-thinking models.

\begin{figure*}[!ht]
    \centering
    \begin{subfigure}[b]{0.24\textwidth}
        \centering
        \includegraphics[width=\textwidth]{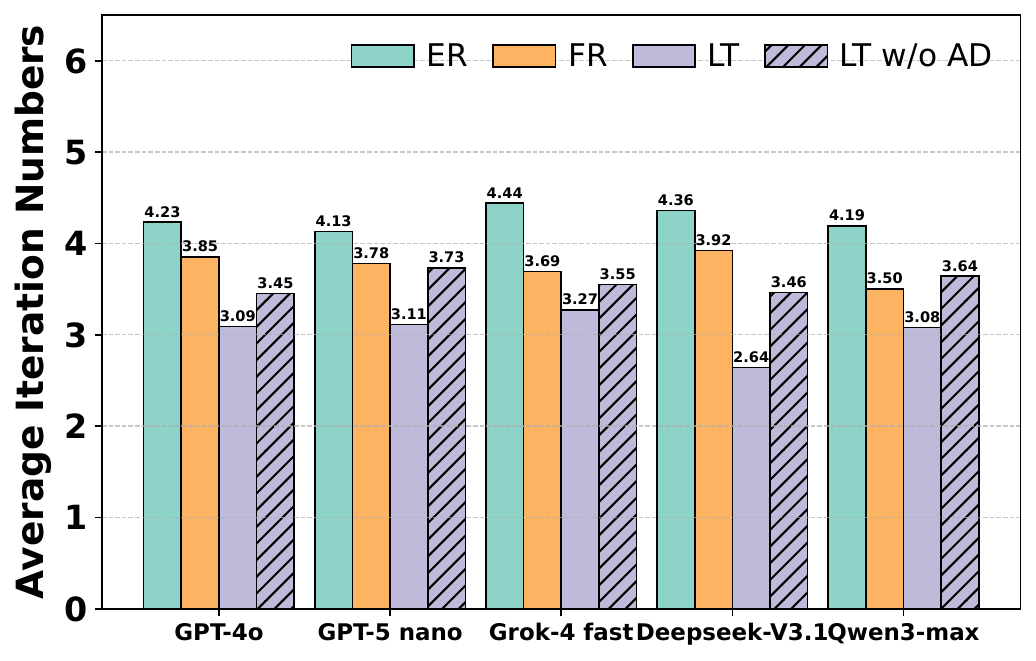}
        \caption{FOLIO}
        \label{fig:folio_iteration}
    \end{subfigure}
    \hfill
    \begin{subfigure}[b]{0.24\textwidth}
        \centering
        \includegraphics[width=\textwidth]{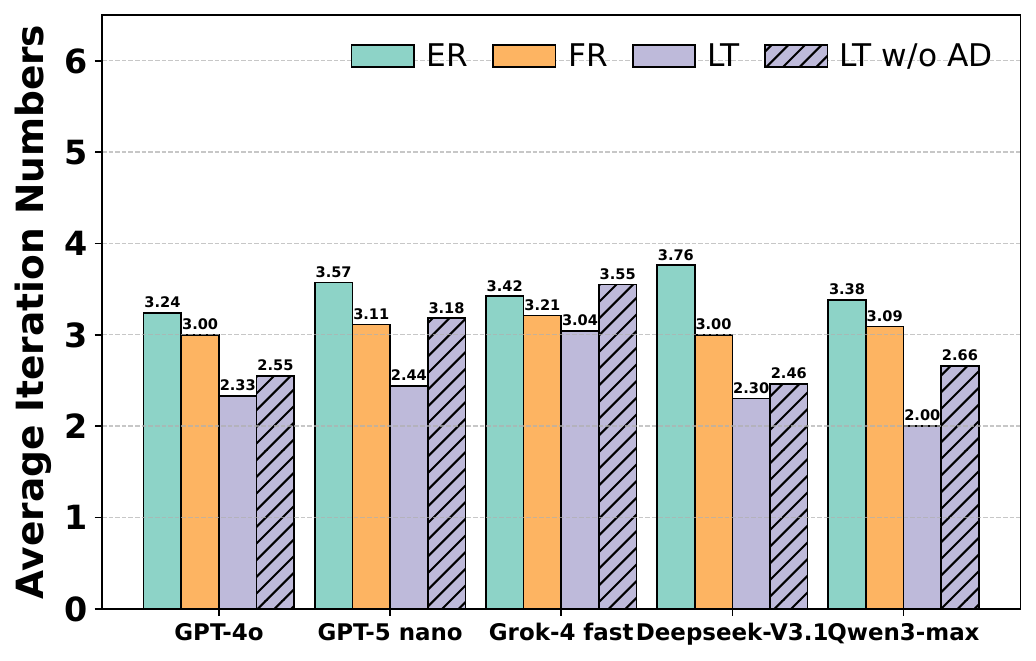}
        \caption{ProofWriter}
        \label{fig:proofwriter_iteration}
    \end{subfigure}
    \hfill
    \begin{subfigure}[b]{0.24\textwidth}
        \centering
        \includegraphics[width=\textwidth]{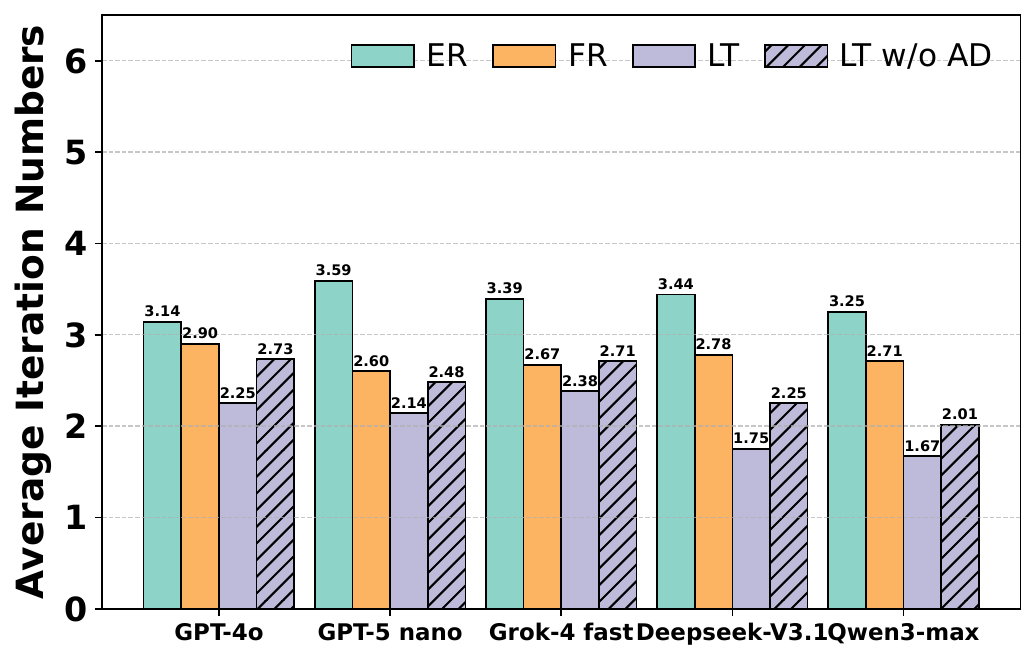}
        \caption{PrOntoQA}
        \label{fig:prontoqa_iteration}
    \end{subfigure}
    \hfill
    \begin{subfigure}[b]{0.24\textwidth}
        \centering
        \includegraphics[width=\textwidth]{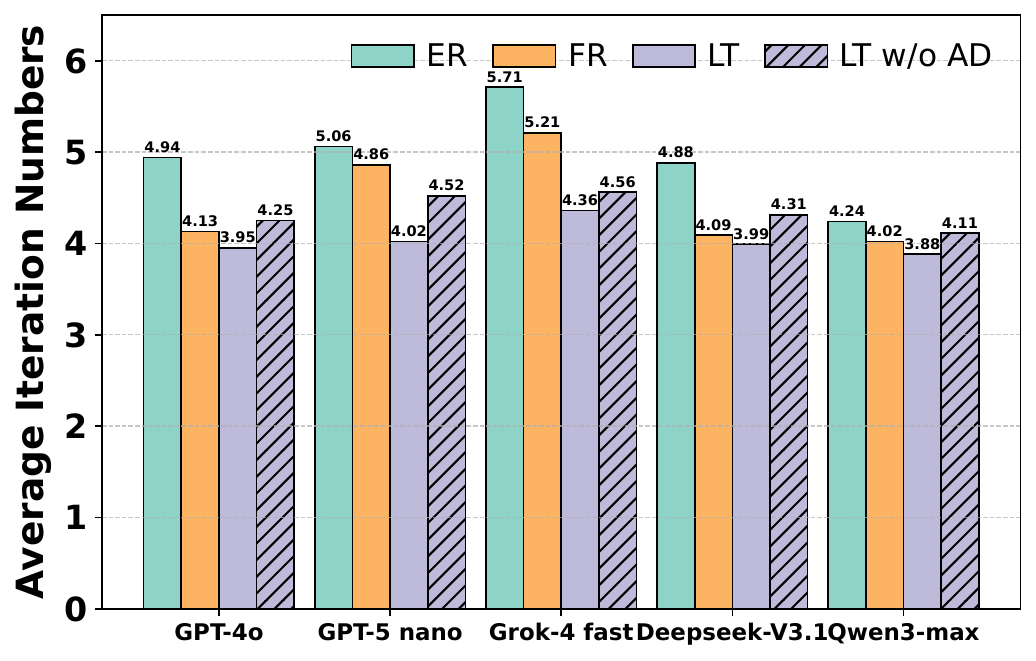}
        \caption{EntailmentBank}
        \label{fig:entailmentbank_iteration}
    \end{subfigure}
    \hfill
     \begin{subfigure}[b]{0.24\textwidth}
        \centering
        \includegraphics[width=\textwidth]{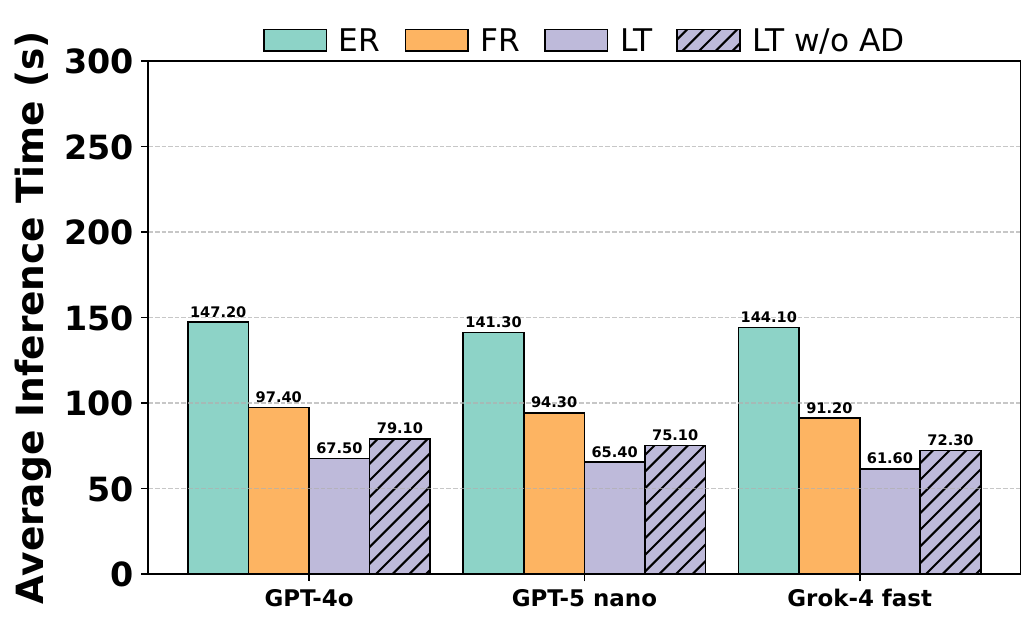}
        \caption{FOLIO}
        \label{fig:folio_inference_time}
    \end{subfigure}
    \hfill
    \begin{subfigure}[b]{0.24\textwidth}
        \centering
        \includegraphics[width=\textwidth]{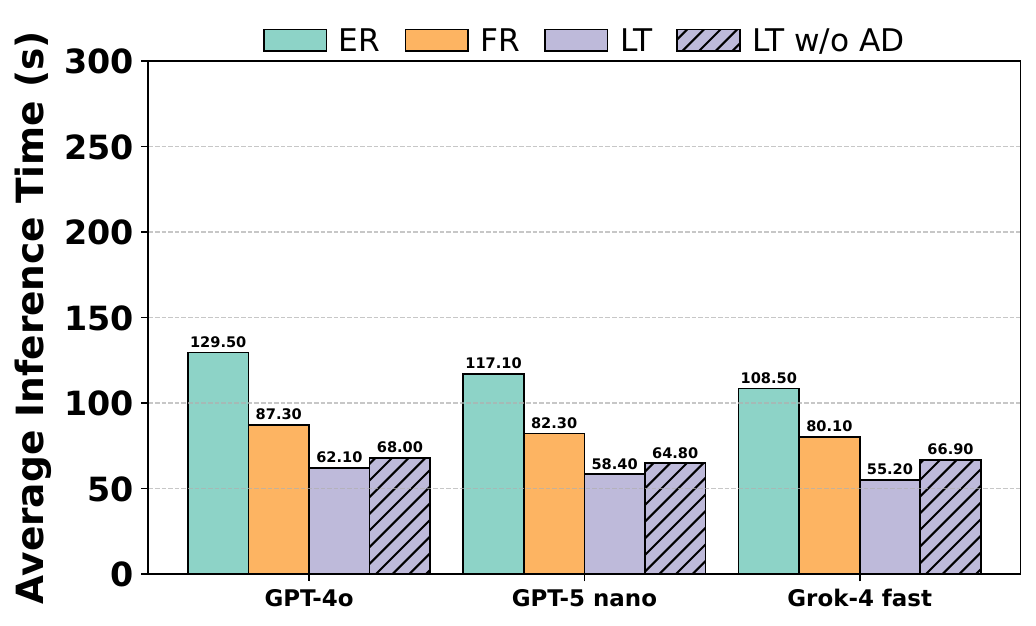}
        \caption{ProofWriter}
        \label{fig:proofwriter_inference_time}
    \end{subfigure}
    \hfill
    \begin{subfigure}[b]{0.24\textwidth}
        \centering
        \includegraphics[width=\textwidth]{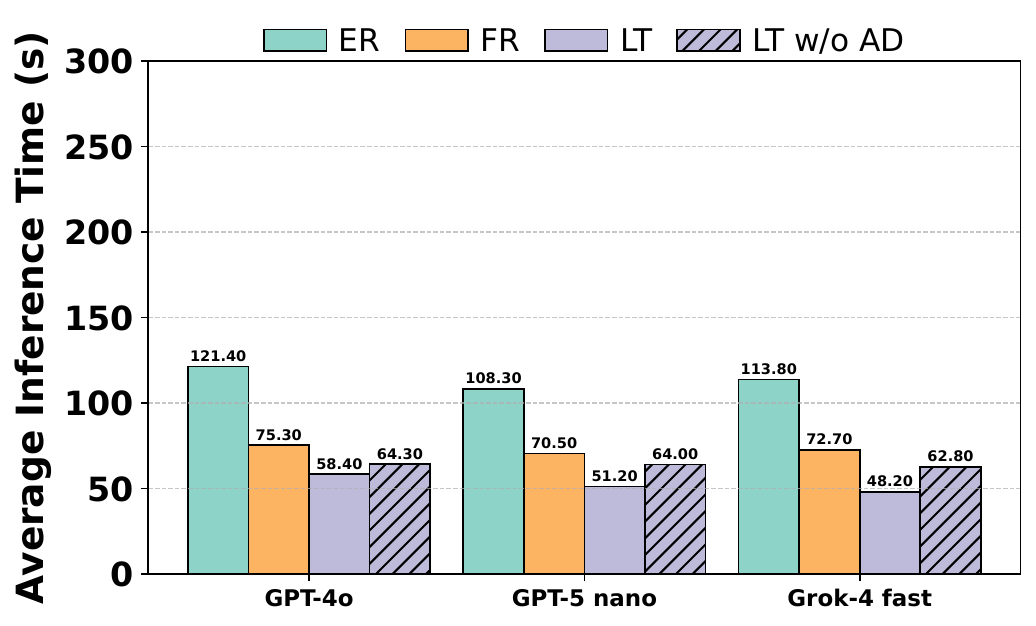}
        \caption{PrOntoQA}
        \label{fig:prontoqa_inference_time}
    \end{subfigure}
    \hfill
    \begin{subfigure}[b]{0.24\textwidth}
        \centering
        \includegraphics[width=\textwidth]{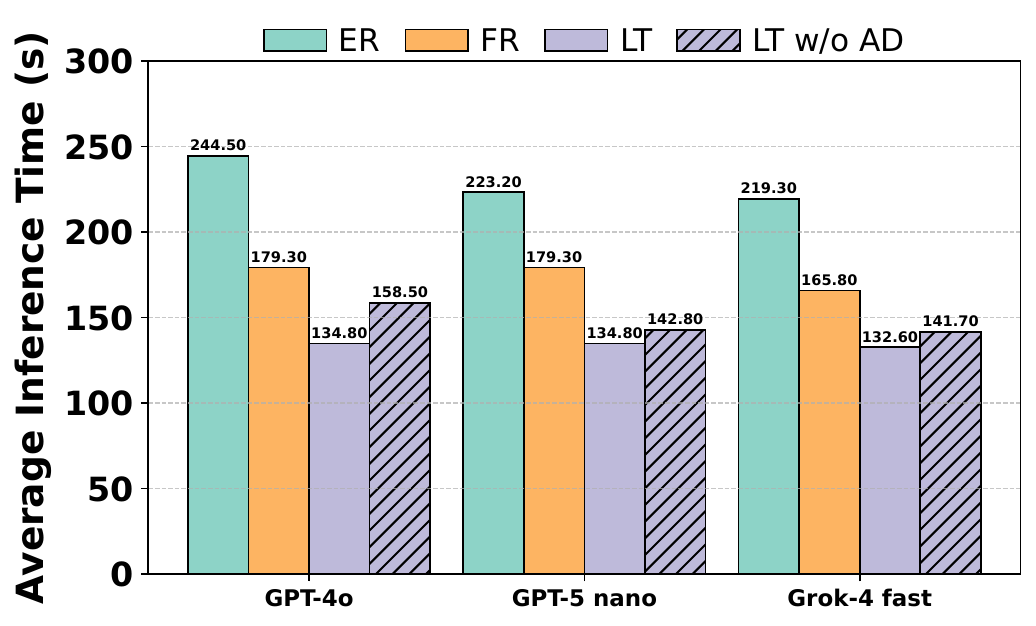}
        \caption{EntailmentBank}
        \label{fig:entailmentbank_inference_time}
    \end{subfigure}
    \caption{Top: Average number of iterations required to refine an explanation. Bottom: Average inference time per iteration. Comparison between ER (Explanation-Refiner), FR (Faithful-Refiner), LT (LLM-TP Tree) and LT w/o AD (LT without atomic decomposition).}
    \label{fig:comparison_efficiency}
\end{figure*}

\begin{figure*}[!ht]
    \centering
     \begin{subfigure}[b]{0.24\textwidth}
        \centering
        \includegraphics[width=\textwidth]{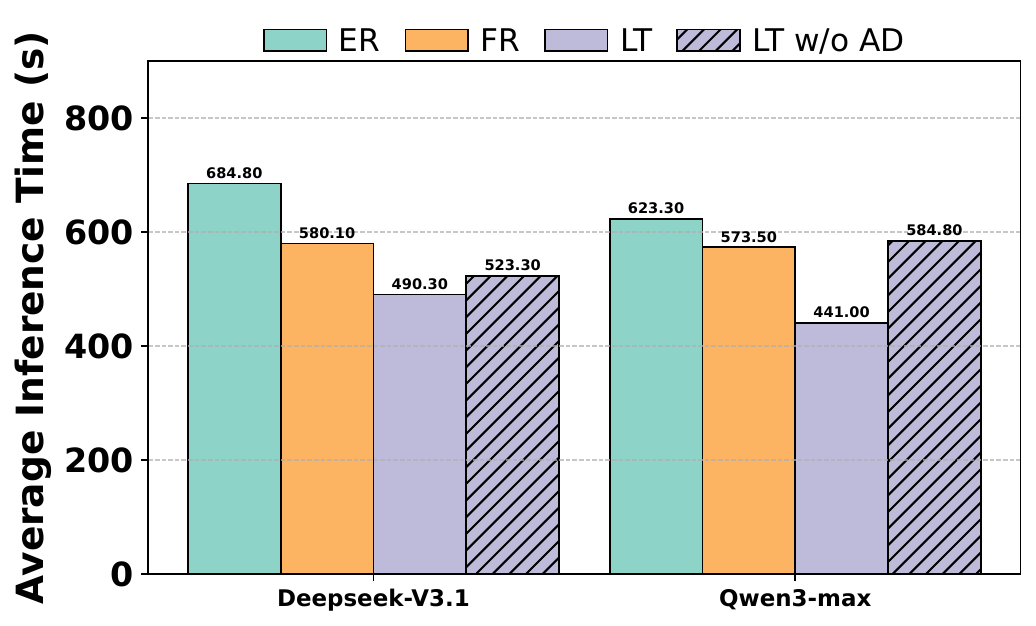}
        \caption{FOLIO}
        \label{fig:folio_inference_time_thinking}
    \end{subfigure}
    \hfill
    \begin{subfigure}[b]{0.24\textwidth}
        \centering
        \includegraphics[width=\textwidth]{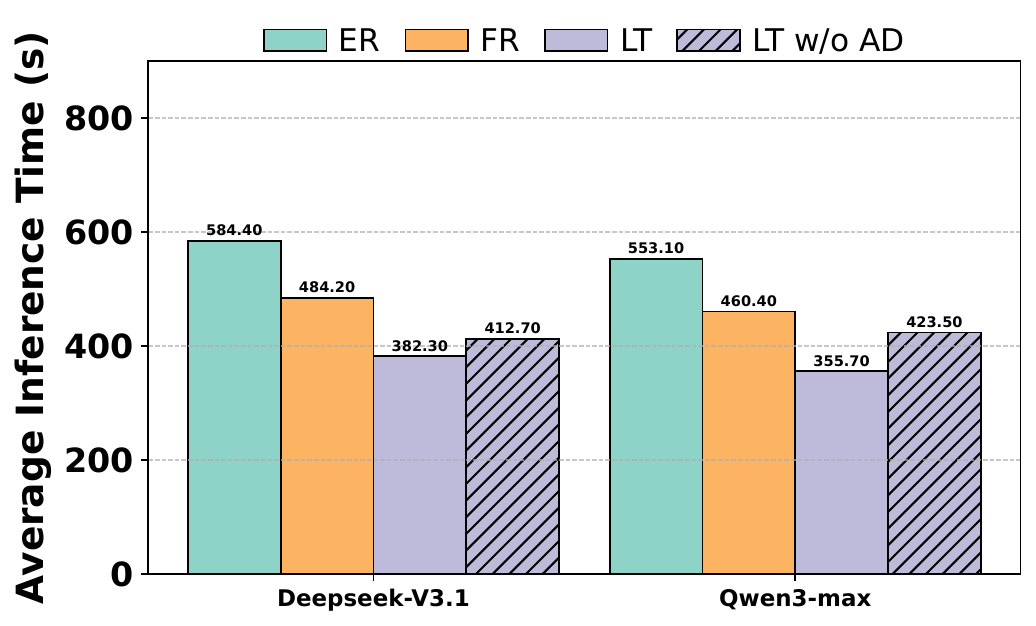}
        \caption{ProofWriter}
        \label{fig:proofwriter_inference_time_thinking}
    \end{subfigure}
    \hfill
    \begin{subfigure}[b]{0.24\textwidth}
        \centering
        \includegraphics[width=\textwidth]{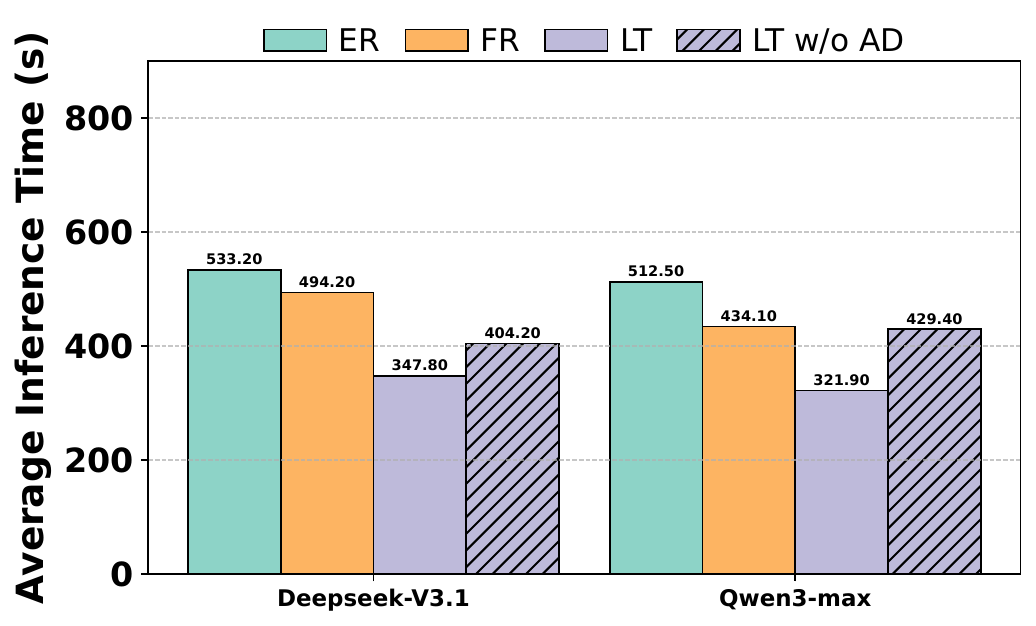}
        \caption{PrOntoQA}
        \label{fig:prontoqa_inference_time_thinking}
    \end{subfigure}
    \hfill
    \begin{subfigure}[b]{0.24\textwidth}
        \centering
        \includegraphics[width=\textwidth]{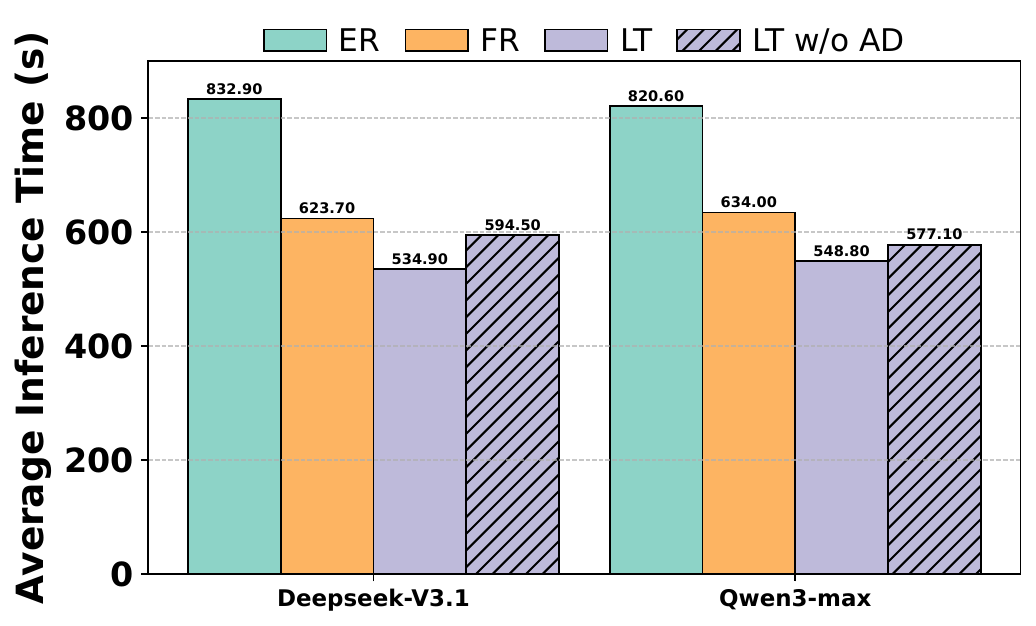}
        \caption{EntailmentBank}
        \label{fig:entailmentbank_inference_time_thinking}
    \end{subfigure}
    \caption{Average inference time per iteration in thinking mode LLMs. Comparison between ER (Explanation-Refiner), FR (Faithful-Refiner), LT (LLM-TP Tree) and LT w/o AD (LT without atomic decomposition).}
    \label{fig:comparison_efficiency_thinking}
\end{figure*}

\section{Full Results: Autoformalisation Faithfulness Across LLM Backbones}
Figure~\ref{fig:comparison_faithfulness} compares the average faithfulness of the autoformalisation process across different approaches, averaged over all LLM backbones.

\begin{figure*}[!ht]
    \centering
    \begin{subfigure}[b]{0.24\textwidth}
        \centering
        \includegraphics[width=\textwidth]{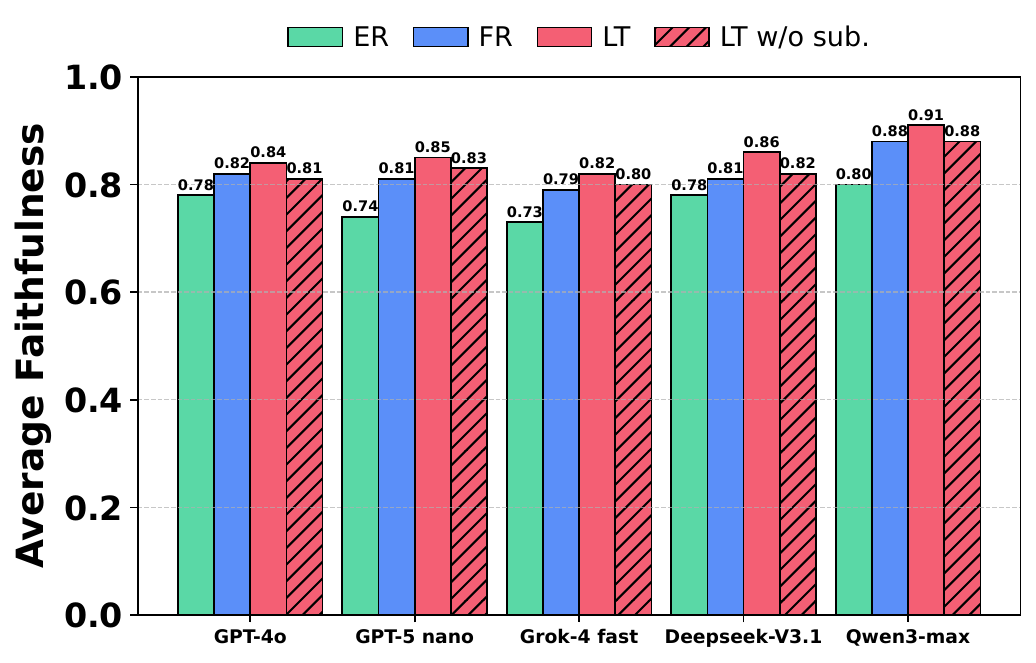}
        \caption{FOLIO}
        \label{fig:folio_faithfulness}
    \end{subfigure}
    \hfill
    \begin{subfigure}[b]{0.24\textwidth}
        \centering
        \includegraphics[width=\textwidth]{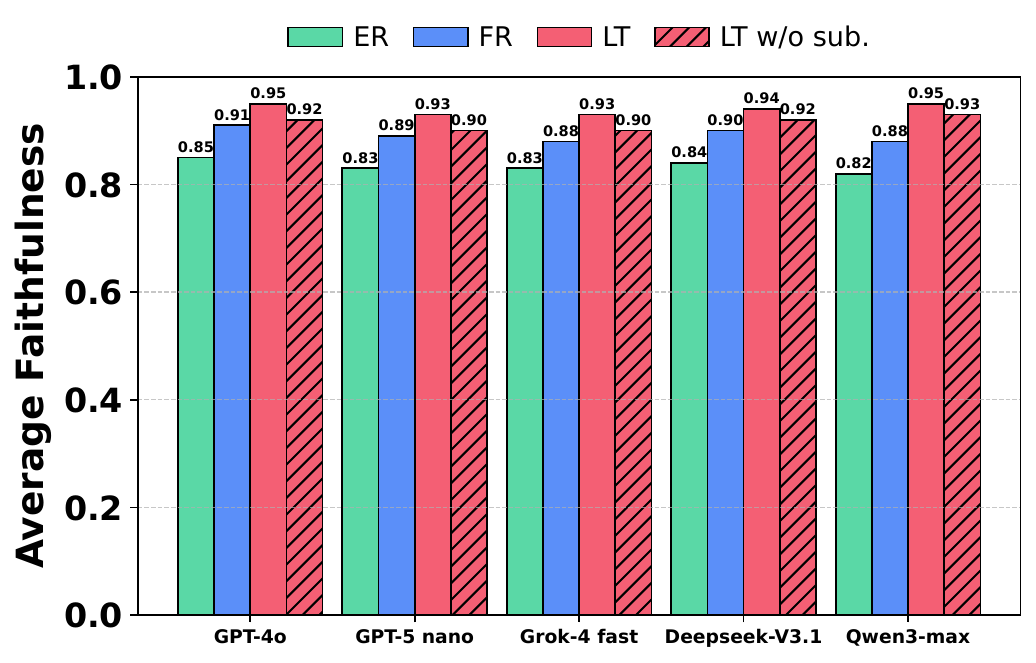}
        \caption{ProofWriter}
        \label{fig:proofwriter_faithfulness}
    \end{subfigure}
    \hfill
    \begin{subfigure}[b]{0.24\textwidth}
        \centering
        \includegraphics[width=\textwidth]{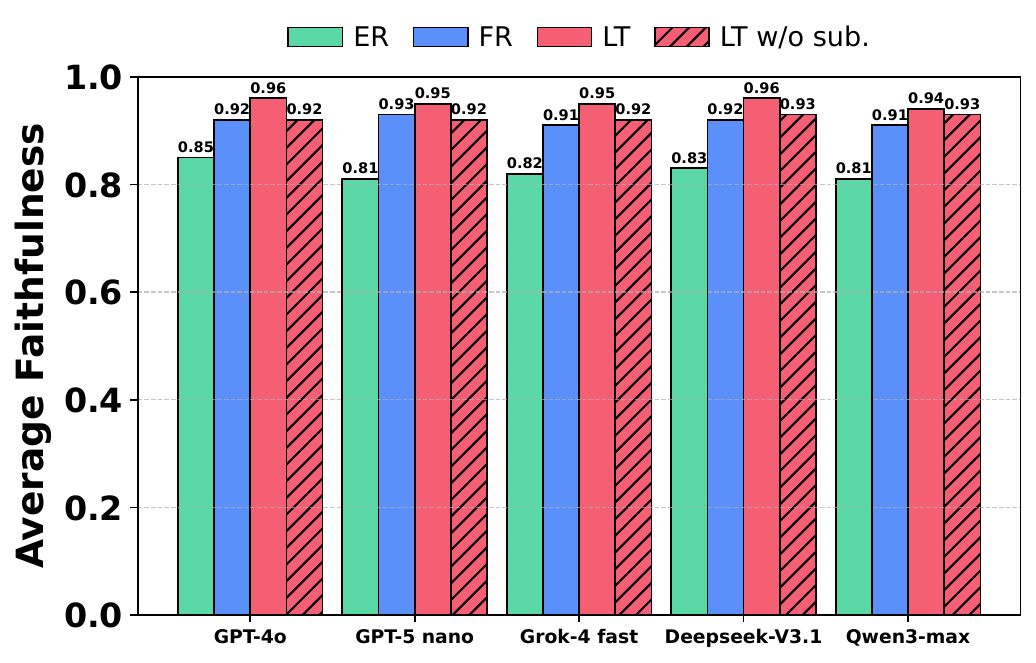}
        \caption{PrOntoQA}
        \label{fig:prontoqa_faithfulness}
    \end{subfigure}
    \hfill
    \begin{subfigure}[b]{0.24\textwidth}
        \centering
        \includegraphics[width=\textwidth]{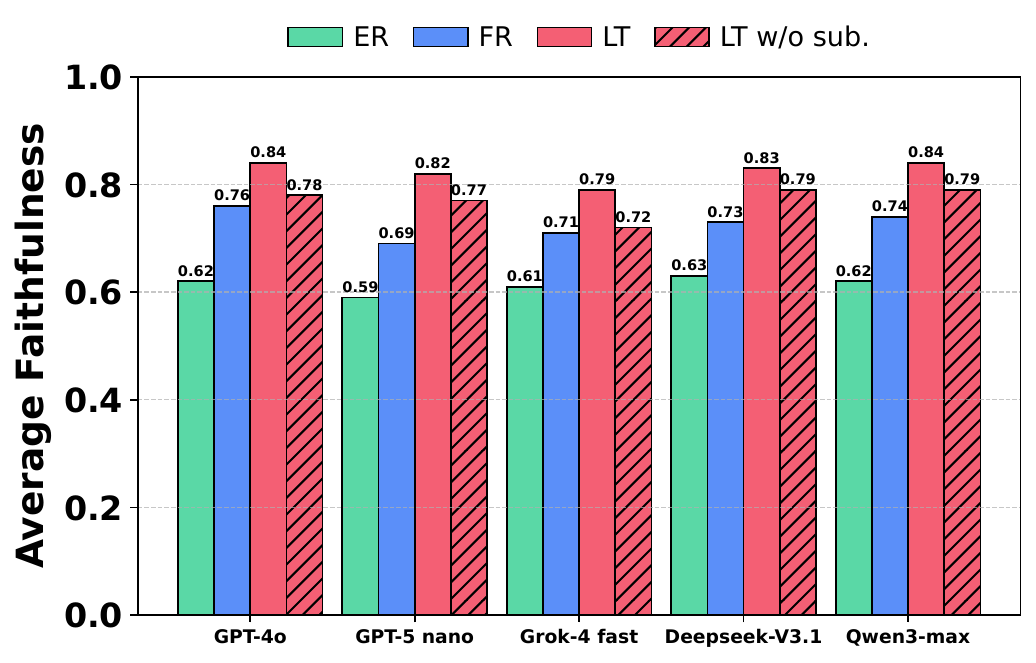}
        \caption{EntailmentBank}
        \label{fig:entailmentbank_faithfulness}
    \end{subfigure}
    \caption{Comparison of the average faithfulness of the autoformalisation process across different approaches.}
    \label{fig:comparison_faithfulness}
\end{figure*}

\section{Solver Proof Depths}
\label{appendix:proof_depth}

Table \ref{tab:depth_alignment_per_llm_mean_refined} and Table \ref{tab:depth_alignment_per_llm_mean_unrefined} shows Per-LLM depth alignment details for ProofWriter and EntailmentBank in refined and unrefined cases. Each cell reports the mean used proof depth averaged over instances within the corresponding gold depth bucket, for a fixed framework and LLM backbone.

% =========================
% Appendix: Per-LLM depth details (mean used depth)
% =========================
\begin{table*}[t]
\centering
\scriptsize
\setlength{\tabcolsep}{3.2pt}
\renewcommand{\arraystretch}{1.15}

\begin{minipage}{0.49\textwidth}
\centering
\textbf{(a) ProofWriter: mean used proof depth by gold depth}\\[2pt]
\resizebox{\linewidth}{!}{%
\begin{tabular}{@{}llccccc@{}}
\toprule
\multirow{2}{*}{\textbf{Framework}} & \multirow{2}{*}{\textbf{LLM}} &
\multicolumn{5}{c}{\textbf{Gold depth $d$}} \\
\cmidrule(lr){3-7}
 &  & 1 & 2 & 3 & 4 & 5 \\
\midrule

\multirow{5}{*}{Explanation-Refiner}
& GPT-4o        & 1.00 & 2.32 & 2.51 & 2.92 & 3.23 \\
& GPT-5 nano    & 1.14 & 2.00 & 2.67 & 2.61 & 2.94 \\
& Grok-4 fast   & 1.08 & 2.04 & 2.53 & 2.62 & 2.92 \\
& Deepseek-V3.1 & 1.28 & 2.41 & 2.23 & 2.87 & 3.04 \\
& Qwen3-max     & 1.10 & 2.43 & 2.24 & 2.90 & 3.11 \\
\midrule

\multirow{5}{*}{Faithful-Refiner}
& GPT-4o        & 1.00 & 2.26 & 2.76 & 3.43 & 3.94 \\
& GPT-5 nano    & 1.02 & 2.04 & 3.13 & 3.34 & 3.63 \\
& Grok-4 fast   & 1.14 & 2.09 & 2.67 & 3.18 & 3.87 \\
& Deepseek-V3.1 & 1.00 & 2.17 & 2.78 & 3.41 & 3.73 \\
& Qwen3-max     & 1.00 & 2.22 & 2.77 & 3.58 & 3.79 \\
\midrule

\multirow{5}{*}{LLM-TP Tree}
& GPT-4o        & 1.00 & 1.99 & 2.94 & 3.76 & 4.72\\
& GPT-5 nano    & 1.08 & 2.12 & 2.98 & 3.68 & 4.53 \\
& Grok-4 fast   & 1.17 & 1.89 & 3.18 & 3.49 & 5.03 \\
& Deepseek-V3.1 & 1.00 & 2.03 & 3.08 & 3.84 & 4.94 \\
& Qwen3-max     & 1.00 & 2.00 & 3.00 & 3.89 & 4.88 \\
\bottomrule
\end{tabular}%
}
\end{minipage}
\hfill
\begin{minipage}{0.49\textwidth}
\centering
\textbf{(b) EntailmentBank: mean used proof depth by gold depth}\\[2pt]
\resizebox{\linewidth}{!}{%
\begin{tabular}{@{}llccccc@{}}
\toprule
\multirow{2}{*}{\textbf{Framework}} & \multirow{2}{*}{\textbf{LLM}} &
\multicolumn{5}{c}{\textbf{Gold Depths $d$}} \\
\cmidrule(lr){3-7}
 &  & 1 & 2 & 3 & 4 & 5 \\
\midrule

\multirow{5}{*}{Explanation-Refiner}
& GPT-4o        & 1.23 & 2.14 & 2.65 & 3.15 & 3.21 \\
& GPT-5 nano    & 1.21 & 2.23 & 2.43 & 2.91 & 2.94 \\
& Grok-4 fast   & 1.13 & 2.21 & 2.52 & 2.93 & 2.92 \\
& Deepseek-V3.1 & 1.16 & 2.28 & 2.67 & 3.04 & 3.16 \\
& Qwen3-max     & 1.28 & 2.32 & 2.63 & 3.18 & 3.20 \\
\midrule

\multirow{5}{*}{Faithful-Refiner}
& GPT-4o        & 1.12 & 2.10 & 2.87 & 3.33 & 3.67 \\
& GPT-5 nano    & 1.11 & 2.12 & 2.64 & 3.22 & 3.40 \\
& Grok-4 fast   & 1.17 & 2.13 & 2.72 & 3.58 & 3.42\\
& Deepseek-V3.1 & 1.03 & 2.25 & 2.78 & 3.40 & 3.69 \\
& Qwen3-max     & 1.17 & 2.20 & 2.93 & 3.47 & 3.73 \\
\midrule

\multirow{5}{*}{LLM-TP Tree}
& GPT-4o        & 1.00 & 2.10 & 2.85 & 3.90 & 4.81 \\
& GPT-5 nano    & 1.00 & 1.84 & 2.84 & 3.86 & 4.71 \\
& Grok-4 fast   & 1.00 & 2.04 & 2.92 & 3.88 & 4.73 \\
& Deepseek-V3.1 & 1.00 & 1.97 & 3.03 & 4.03 & 4.85 \\
& Qwen3-max     & 1.00 & 1.96 & 2.94 & 3.98 & 4.92 \\
\bottomrule
\end{tabular}%
}
\end{minipage}
\caption{Per-LLM depth alignment details for ProofWriter and EntailmentBank in refined cases.}
\label{tab:depth_alignment_per_llm_mean_refined}
\end{table*}

\begin{table*}[t]
\centering
\scriptsize
\setlength{\tabcolsep}{3.2pt}
\renewcommand{\arraystretch}{1.15}

\begin{minipage}{0.49\textwidth}
\centering
\textbf{(a) ProofWriter: mean used proof depth by gold depth}\\[2pt]
\resizebox{\linewidth}{!}{%
\begin{tabular}{@{}llccccc@{}}
\toprule
\multirow{2}{*}{\textbf{Framework}} & \multirow{2}{*}{\textbf{LLM}} &
\multicolumn{5}{c}{\textbf{Gold depth $d$}} \\
\cmidrule(lr){3-7}
 &  & 1 & 2 & 3 & 4 & 5 \\
\midrule

\multirow{5}{*}{Explanation-Refiner}
& GPT-4o        & 1.00 & 2.34 & 2.49 & 2.57 & 2.33 \\
& GPT-5 nano    & 1.12 & 1.99 & 2.57 & 2.46 & 2.24 \\
& Grok-4 fast   & 1.10 & 1.87 & 2.48 & 2.38 & 2.32 \\
& Deepseek-V3.1 & 1.18 & 2.37 & 1.98 & 2.84 & 2.54 \\
& Qwen3-max     & 1.00 & 2.33 & 2.11 & 2.47 & 2.62 \\
\midrule

\multirow{5}{*}{Faithful-Refiner}
& GPT-4o        & 1.02 & 2.16 & 2.73 & 3.44 & 3.12 \\
& GPT-5 nano    & 1.02 & 1.95 & 3.23 & 3.24 & 3.02 \\
& Grok-4 fast   & 1.16 & 2.46 & 2.62 & 3.09 & 3.06 \\
& Deepseek-V3.1 & 1.00 & 1.98 & 2.78 & 2.98 & 3.27 \\
& Qwen3-max     & 1.00 & 2.22 & 2.67 & 3.15 & 3.43 \\
\midrule

\multirow{5}{*}{LLM-TP Tree}
& GPT-4o        & 1.02 & 2.13 & 3.01 & 3.72 &  4.72\\
& GPT-5 nano    & 1.09 & 2.14 & 2.95 & 3.54 & 4.65 \\
& Grok-4 fast   & 1.15 & 1.79 & 3.16 & 3.50 & 5.13 \\
& Deepseek-V3.1 & 1.02 & 2.13 & 3.04 & 3.86 & 4.87 \\
& Qwen3-max     & 1.00 & 1.98 & 3.00 & 3.88 & 4.91 \\
\bottomrule
\end{tabular}%
}
\end{minipage}
\hfill
\begin{minipage}{0.49\textwidth}
\centering
\textbf{(b) EntailmentBank: mean used proof depth by gold depth}\\[2pt]
\resizebox{\linewidth}{!}{%
\begin{tabular}{@{}llccccc@{}}
\toprule
\multirow{2}{*}{\textbf{Framework}} & \multirow{2}{*}{\textbf{LLM}} &
\multicolumn{5}{c}{\textbf{Gold Depths $d$}} \\
\cmidrule(lr){3-7}
 &  & 1 & 2 & 3 & 4 & 5 \\
\midrule

\multirow{5}{*}{Explanation-Refiner}
& GPT-4o        & 1.53 & 2.84 & 2.48 & 2.44 & 2.21 \\
& GPT-5 nano    & 1.39 & 2.99 & 2.33 & 2.31 & 1.95 \\
& Grok-4 fast   & 1.43 & 2.83 & 2.24 & 2.21 & 1.92 \\
& Deepseek-V3.1 & 1.48 & 2.58 & 2.33 & 2.14 & 1.96 \\
& Qwen3-max     & 1.88 & 3.12 & 2.32 & 2.19 & 2.12 \\
\midrule

\multirow{5}{*}{Faithful-Refiner}
& GPT-4o        & 1.34 & 2.42 & 2.48 & 3.23 & 3.55 \\
& GPT-5 nano    & 1.33 & 2.32 & 2.41 & 3.32 & 3.30 \\
& Grok-4 fast   & 1.49 & 2.28 & 2.43 & 3.54 & 3.42\\
& Deepseek-V3.1 & 1.10 & 2.85 & 2.34 & 3.38 & 3.45 \\
& Qwen3-max     & 1.67 & 2.74 & 2.53 & 3.43 & 3.42 \\
\midrule

\multirow{5}{*}{LLM-TP Tree}
& GPT-4o        & 1.03 & 2.13 & 2.83 & 3.93 & 4.62 \\
& GPT-5 nano    & 1.13 & 1.86 & 2.80 & 3.78 & 4.51 \\
& Grok-4 fast   & 1.12 & 2.04 & 2.95 & 3.91 & 4.43 \\
& Deepseek-V3.1 & 1.03 & 1.89 & 3.12 & 4.12 & 4.67 \\
& Qwen3-max     & 1.00 & 2.01 & 2.99 & 4.08 & 4.74 \\
\bottomrule
\end{tabular}%
}
\end{minipage}
\caption{Per-LLM depth alignment details for ProofWriter and EntailmentBank in unrefined cases.}
\label{tab:depth_alignment_per_llm_mean_unrefined}
\end{table*}

\section{Atomic Decomposition Prompt}
\label{appendix:atomic_decompistion_prompt}
We build the atomic decomposition prompt based on \citet{srikanth-rudinger-2025-nli} as: 
\begin{tcolorbox}[width=0.5\textwidth, 
    before skip=2pt,   % 
    after skip=0pt,    % 
    top=3pt,          % 
    bottom=3pt,
    left=3pt,
    right=3pt,
    breakable
    ]
{\small
\textbf{SYSTEM:} You are an expert in symbolic/logical reasoning, linguistic and natural language inference. You are given a sentence and its logical information. Generate a list of atomic facts that are strictly logically entailed from the given sentence.

\textbf{Instructions:}
\begin{enumerate}
    \item Keep each fact independent and self-contained.
    \item Each fact should make sense when read on its own.
    \item Only write facts that are directly described or supported by the sentence.
    \item The atomic facts must be logically entailed from the given sentence.
\end{enumerate}

\textbf{USER:} Here are some examples:

\textbf{Example:}\\
\textbf{Provided Sentence:}\\
Professional actors are in a summer performance.

\textbf{Answer:}\\
Atom 1: The people are professional.\\
Atom 2: The people are actors.\\
Atom 3: The performance is during summer.

\textbf{Task:}\\
\textbf{Provided Sentence:} \texttt{\{sentence\}}\\

\textbf{Answer:}
}
\end{tcolorbox}

\section{Entailment Tree Construction Prompt}
\label{appendix:entailment_tree}
\begin{tcolorbox}[width=0.5\textwidth, 
    before skip=2pt,   % 
    after skip=0pt,    % 
    top=3pt,          % 
    bottom=3pt,
    left=3pt,
    right=3pt,
    breakable
    ]
{\small
\textbf{SYSTEM:} You are an expert in natural language inference and textual entailment. Given the following premise sentences and a final conclusion, generate some step-by-step intermediate conclusion sentences to finally infer the final conclusion.\\
\textbf{Instructions:}
\begin{enumerate}
    \item The intermediate conclusion sentences must be strictly logical entailed from the premise sentences. 
    \item Since the reasoning is step-by-step, the intermediate conclusion sentences can be the new premise sentences for next step.
    \item One intermediate conclusion sentence can be generated from multiple premise sentences.
    \item If the final conclusion can be directly inferred from the premise sentences, you need to state the intermediate conclusion sentences are empty.
    \item There might be redundant premise sentences.
\end{enumerate}

\textbf{USER:} Here are some examples:

\textbf{Example:}\\
Initial Premises:\\
1. Monkeypox is an infectious disease caused by the monkeypox virus.\\
2. Monkeypox virus can occur in certain animals, including humans.\\
3. Humans are mammals.\\
4. Mammals are animals.\\
5. Symptoms of Monkeypox include fever, headache, muscle pains, feeling tired, and so on.\\
6. People feel tired when they get a flu.\\
\\
Final Conclusion:\\
There is an animal.\\
\\
Answer:\\
Monkeypox is an infectious disease caused by the monkeypox virus.\\
Monkeypox virus can occur in certain animals, including humans.\\
Conclusion: Humans can get monkeypox.\\
\\
Humans are mammals.\\
Mammals are animals.\\
Conclusion:Humans are animals.\\
\\
...
\\
Humans are animals.\\
Therefore, there is an animal (humans) that can get monkeypox and feel tired.
Conclusion:Therefore, there is an animal.\\
\\
Initial Premises:\\\\
\\
Final Conclusion:\\\\
\\
Answer:\\
}
\end{tcolorbox}

\section{$\theta$-substitution Autoformalisation Prompt}
\begin{tcolorbox}[width=0.5\textwidth, 
    before skip=2pt,   % 
    after skip=0pt,    % 
    top=3pt,          % 
    bottom=3pt,
    left=3pt,
    right=3pt,
    breakable
    ]
{\small
\textbf{SYSTEM:} You are an expert in symbolic/logical reasoning and autoformalisation. You are given a sentence.Extract the logical relation from a given natural language sentence and construct a logical template representing its structure, following the format demonstrated in the examples.\\ 

\textbf{Instructions:}
\begin{enumerate}
    \item Read and understand the provided sentence.
    \item Identify key entities, relationships, and logical structure (e.g., quantifiers, conjunctions, disjunctions, implications).
    \item Abstract the sentence into a logical template using placeholders (e.g., $\forall$ x y z. P(x) $\wedge$ Q(y) $\wedge$ R(z) $\rightarrow$ S) that capture the logical form, not the specific content.
\end{enumerate}
Follow these steps:\\
1. Parse the sentence and reason step-by-step about its logical components: What are its predicates, quantifiers, variables, and connectives? \\
2. Formulate the logical template, using generic predicate letters (P, Q, R, S, etc.) and variable placeholders (x, y, z, etc.), mirroring the schema shown in the examples.\\
...
\\
Example 1:\\
Sentence: If someone wins the lottery, they will buy a new house or a car.\\
\\
Example 1 Answer:\\
Logical Template: $\forall$x y z. P(x) $\wedge$ Q(y) $\rightarrow$ (R(z) $\vee $S(z))\\
...\\
\textbf{Provided Sentence:} \texttt{\{sentence\}}\\
\\
\textbf{Answer:}
}
\end{tcolorbox}

\end{document}